\documentclass[10pt,journal]{IEEEtran}
\ifCLASSOPTIONcompsoc
  \usepackage[nocompress]{cite}
\else
  \usepackage{cite}
\fi
\ifCLASSINFOpdf
\else
\fi
\newcommand{\secref}[1]{Section~\ref{sec:#1}}
\newcommand{\figref}[1]{Figure~\ref{fig:#1}}

\newcommand{\tabref}[1]{Table~\ref{tab:#1}}
\newcommand{\eqnref}[1]{Equation~\ref{equ:#1}}
\newcommand{\eqnoref}[1]{(\ref{equ:#1})}
\newcommand{\etal}{\mbox{\emph{et al.\ }}}
\usepackage[numbers]{natbib}
\usepackage[caption=false,font=footnotesize]{subfig}
\usepackage[cmex10]{amsmath}
\usepackage{amssymb}
\usepackage{algorithm}
\usepackage{algorithmic}
\usepackage{multirow}

\usepackage{color,soul}
\usepackage[framemethod=tikz]{mdframed}

\begin{document}
%
\title{Effective Sampling: Fast Segmentation Using Robust Geometric Model Fitting}
%
%
%
%

\author{Ruwan~Tennakoon,
        Alireza~Sadri,
        Reza~Hoseinnezhad,
        and~Alireza~Bab-Hadiashar,~\IEEEmembership{Senior~Member,~IEEE}
\IEEEcompsocitemizethanks{\IEEEcompsocthanksitem R.B. Tennakoon, A. Sadri, R. Hoseinnezhad and A. Bab-Hadiashar are with the School of Engineering, RMIT University, Melbourne,
Australia.\protect\\
E-mail: ruwan.tennakoon@rmit.edu.au}%
}

\IEEEtitleabstractindextext{%
\begin{abstract}
Identifying the underlying models in a set of data points contaminated by noise and outliers, leads to a highly complex multi-model fitting problem. This problem can be posed as a clustering problem by the projection of higher order affinities between data points into a graph, which can then be clustered using spectral clustering. Calculating all possible higher order affinities is computationally expensive. Hence in most cases only a subset is used.
In this paper, we propose an effective sampling method to obtain a highly accurate approximation of the full graph required to solve multi-structural model fitting problems in computer vision. The proposed method is based on the observation that the usefulness of a graph for segmentation improves as the distribution of hypotheses (used to build the graph) approaches the distribution of actual parameters for the given data. In this paper, we approximate this actual parameter distribution using a k-th order statistics based cost function and the samples are generated using a greedy algorithm coupled with a data sub-sampling strategy. 
The experimental analysis shows that the proposed method is both accurate and computationally efficient compared to the state-of-the-art robust multi-model fitting techniques. The code is publicly available from https://github.com/RuwanT/model-fitting-cbs.
\end{abstract}

}

\maketitle

\IEEEdisplaynontitleabstractindextext

%
\IEEEpeerreviewmaketitle

\IEEEraisesectionheading{\section{Introduction}\label{sec:introduction}}
Robust fitting of geometric models to data contaminated with both noise and outliers is a well studied problem with many applications in computer vision \cite{Fischler1981, Delong2012, Elhamifar2013, Haifeng2003}. Visual data often contain multiple underlying structures and there are pseudo-outliers (measurements representing structured other than the structure of interest \cite{Stewart1997}) as well as gross-outliers (produced by errors in the data generation process). Fitting models to this combination of data involves solving a highly complex multi-model fitting problem. The above multi-model fitting problem can be viewed as a combination of two sub problems: \textit{data labeling} and \textit{model estimation}. Although solving one of the sub-problems, when the solution to the other is given, is straightforward, solving both problems simultaneously remains a challenge.

Traditional approaches to multi-model fitting were based on fit and remove strategy: apply a high breakdown robust estimator (e.g. RANSAC \cite{Fischler1981}, least k-th order residual) to generate a model estimate and remove its inliers to prevent the estimator from converging to the same structure again. However, this approach is not optimal as errors made in the initial stages tend to make the subsequent steps unreliable (e.g. small structures can be absorbed by models that are created by accidental alignment of outliers with several structures) \cite{Zuliani2005}. To address this issue, energy minimization methods have been proposed. They are based on optimizing a cost function consisting of a combination of data fidelity and model complexity (number of model instances) terms \cite{Boykov2001}. In this approach, the cost function is optimized to simultaneously recover the number of structures and their data association. Commonly such cost functions are optimized using discrete optimization methods (metric labeling \cite{Delong2012}). They start form a large number of proposal hypotheses and gradually converge to the true models. The outcome of those methods depends on the appropriate balance between the two terms in the cost function (controlled by an input parameter) as well as the quality of initial hypotheses. The method proposed in this paper is primarily designed to avoid the use of parameters that are difficult to tune. 
Sensitivity to the parameters included for the summation of terms with different dimensions is also an issue associated with the application of several other subspace learning and clustering methods. For instance, Robust-PCA \cite{candes2011robust} splits the data matrix into a low-rank matrix and a sparse error matrix. The aim is to minimize the cost function (which is a norm of the error matrix) while it is regularized by a rank of representation matrix. In factorization methods such as \cite{cabral2013unifying} the low-rank representation is obtained by learning a dictionary and coefficients for each data point. The effect of regularization is included using a parameter. These parameters often depend on noise scales, complexity of structures and even depend on the number of underlying structures and their data points. As such, these variables vary between data-sets and therefore limits the application of those methods.

Another approach to multi-model fitting is to pose the problem as a clustering problem \cite{Agarwal2005} \cite{Govindu2005}. In this approach, the idea is that a pure sample (members of the same structure) of the observed data from a cluster can be represented by a linear combination of other data points from the same cluster. Then the relations of all points to each sampled subset can encode the relations between data points. For example Sparse Subspace Clustering SSC \cite{Elhamifar2013} tries to find a sparse block-diagonal matrix that relates data points in each cluster. The optimization task in this work is to minimize the error as well as the $L_1$ norm of this latent sparse matrix. In contrast, the regularization term in LRR \cite{LRRliu2013robust} uses nuclear norm of this sparse matrix. Our proposed method is computationally faster than these methods and does not need the parameter brought in both cases for the regularization. Recently \cite{LRRDetliu2016deterministic} gave a deterministic analysis of LRR and suggested that the regularization parameter can be estimated by looking at the number of data points. Although this improves the speed and accuracy of those methods, it remains unclear what would happen when the number of data points is very high (similar to databases studied in this work). We should also note that methods such as LRSR \cite{LRSRNCwang2016lrsr} and CLUSTEN \cite{TIP16kim2016robust}, with more constraints for the regularization and therefore more parameters, have also been proposed. A similar strategy is also taken to solve the problem of Global Dimension Minimization in \cite{poling2014new} which is used to estimate the fundamental matrix for the problem of two-view motion segmentation. The method is somewhat more accurate than LRR and SSC but it is computationally expensive.

Another widely used clustering method is called Spectral Clustering \cite{Ng2002}. The main idea is to search for possible relations between data points and form a graph that encodes the relations obtained by this search. Spectral clustering, based on eigen-analysis of a pairwise similarity graph, finds a partitioning of the similarity graph such that the data points between different clusters have very low similarities and the data points within a cluster have high similarities.  A simple measure of similarity between a pair of points lying on a vector field is the euclidean distance. However, such measures based on just two points will not work when the problem is to identify data points that are explained by a known structure with multiple degrees of freedom. For instance, in a 2D line fitting problem, any two points will perfectly fit a line irrespective of their underlying structure, hence a similarity cannot be derived by just using two points. In such cases an effective similarity measure can be devised using higher order affinities (e.g. for a 2D line fitting problem least square error between three or more points will provide a suitable affinity measure indicating how well those points approximate a line \cite{Agarwal2005}).

There are several methods to represent higher order affinities using either a hyper-graph or a higher order tensor.  Since spectral clustering cannot be applied directly to those higher order representations, they are commonly projected to a graph (discussed further in \secref{background}). It is also known that the number of elements in a higher order affinity tensor (or number of edges in a hyper-graph) will increase exponentially with the order of the affinities ($h$), which is directly related to the complexity of the model ($p$). Hence, for complex models it would not be computationally feasible (in terms of memory utilization or computation time) to generate the full affinity tensor (or hyper-graph) even for a moderate size dataset. The commonly used method to overcome this problem is to use a sampled version of the full tensor (or hyper-graph) obtained using random sampling \cite{Govindu2005}, \cite{Agarwal2005}.  
The information content of the projected graph heavily dependents on the quality of the samples used \cite{Chen2009}, \cite{Ochs2012}, \cite{Purkait2014} and we analyze this behavior in \secref{background}.

In this paper, we propose an efficient sampling method called cost based sampling (CBS), to obtain a highly accurate approximation of the full graph required to solve multi-structural model fitting problems in computer vision. The proposed method is based on the observation that the usefulness of a graph for segmentation improves as the distribution of hypotheses (used to build the graph) approaches the actual parameter distribution for the given data. 
The approach is similar to the one proposed in \cite{MoGCVPRli2015subspace} where Mixture of Gaussian is used to find the structures in the parameter space. The search is initialized by a few Gaussians and the parameters of the mixture is obtained through Expectation-Maximization steps. The grouping strategy is based on the above mentioned optimization approach and similarly involves the use of a regularization parameter that is difficult to tune. When the number of Gaussians is too low, which is to seek a few perfect samples, the noise cannot be characterized properly and some structures may be missed. Increasing the number of Gaussians is computationally expensive for the EM part. This is where our approach is most effective. Our proposed method benefits from a fast greedy optimization method to generate many samples and makes use on the inherent robustness of Spectral Clustering for occasional samples that may not be perfect. 

The underlying assumption in this approach is that the parameter distribution can reveal the underlying structures and the generation of many good samples is the key to properly construct the distribution for successful clustering. This basic approach can be implemented with different choices of cost functions and optimization methods. The choice of the optimization method mostly determines the speed and the choice of the cost function affects the accuracy. For example, LBF \cite{LBFzhang2012hybrid} attempts to improve the generated samples of the cost function (chosen to be the $\beta$-number of the residuals of a model) by guiding the samples and increasing their size. Its optimization method is slower than our proposed method, which uses the derivatives of the cost function and the chosen cost function is very steep around the structures, which makes the initialization of the method very difficult and can lead to missing structures. The recipe to overcome these shortcomings is based on using extra constraint, such as spatial contiguity, to ensure the purity of samples before increasing their sizes. In this paper, we approximate this actual parameter distribution using the k-th order cost function, which in turn enables us to generate samples using a greedy algorithm that incorporates a faster optimization method. The advantage of the proposed method is that it only uses information present in data with respect to a putative model and does not require any additional assumptions such as spatial smoothness. 

The main contribution of this paper is the introduction of a fast and accurate data segmentation method based on effective combination of the accuracy of a new sampling method with the speed of a good clustering method. The paper presents a reformulation of these methods in way that it makes them complementary. The proposed sampler is ensured to visit all structures in data (by a high probability) and guide each sample to represent the closest structure. This is achieved by focusing on the distribution of putative models in parameter space and by providing samples with highest likelihoods from each structure. The choice of maximum likelihood method plays an important role in the speed of the sampler where the accuracy is still preserved. Furthermore, compared to other techniques, the proposed method incorporates less sensitive parameters that are difficult to tune. In particular, we compare the proposed method with ones using a scale parameter to combine two unrelated cost functions. Such a parameter is often data dependent and difficult to tune for a general solution.

The rest of this paper is organized as follows. \secref{background} discusses the use of clustering techniques for robust model fitting and the need for better sampling methods. \secref{method} describes the proposed method in detail and \secref{experimentalanalysis} presents experimental results involving real data, and comparisons with state-of-the-art model-fitting techniques. Additional discussion regarding the merits and shortcomings of the method is presented in \secref{Discussion} followed by a conclusion in \secref{conclusion}.

\section{Background}
\label{sec:background}
Consider the problem of clustering data points $ X = \left[ x_i \right]^N_{i=1}; x_i \in \mathbb{R}^d$ assuming that there are underlying models (structures) $\Theta = \left[ \theta^{(j)}\right]_{j=1}^m; \theta^{(j)} \in \mathbb{R}^p$ that relate some of those points together. Here $N$ is the number of data points and $m$ is the number of structures in the dataset with zeroth structure assigned for outliers. Clustering a data-set, in such a way that elements of the same group have higher similarity than the elements in different groups is a well-studied problem with attractive solutions like spectral clustering. Spectral clustering operates on a pairwise undirected graph with affinity matrix, $G$, that contain affinities between pairs of points in the dataset. As explained earlier, for model fitting applications, only higher than pairwise order affinities reveal useful similarity measure and spectral clustering cannot be directly applied to higher order affinities.

Agrawal \etal \cite{Agarwal2005} introduced an algorithm where the higher order affinities (in multi-structural multi-model fitting problems) were represented as a hyper-graph. They proposed a two step approach to partition a hyper-graph with $h=p+1$ ($p$ is the number of parameters of the model) affinities. In the first step, the hyper-graph was  approximated with a weighted graph using clique averaging technique. The resulting graph was then segmented using spectral clustering. Constructing the hyper-graph with all possible $p+1$ edges is very expensive to implement. As such, they used a sampled version of the hyper-graph constructed by random sampling. 

Govindu \cite{Govindu2005} posed the same problem in a tensor theoretic approach where the higher order affinities were represented as an $h$-dimensional tensor $\mathcal{P}$. Using the relationship between higher order SVD (HOSVD) of the $h$-mode representation and the eigan value decomposition \cite{Govindu2005} showed that the supper symmetric tensor $\mathcal{P}$ (the similarity does not depend on the ordering of points in the $h$-tuple) can be decomposed in to a pairwise affinity matrix using $G = PP^\top$. Here $P$ is the flattened matrix representation\footnote{The flattened matrix ($P_d$) along dimension $d$ is a matrix with each column obtained by varying the index along dimension $d$ while holding all other dimensions fixed.} of $\mathcal{P}$ along any dimension. The size of the matrix $P$ is still very large. For example, the size of $P$ for a similarity tensor constructed using $h$-tuples from a dataset containing $N$ data points is $N \times N^{h-1}$. As with the hyper-graphs, to make the computation tractable  Govindu \cite{Govindu2005} suggested a sampled version of the flattened matrix ($H \approx P$) to be used. Each column of $H$ was obtained using the residuals to a model ($\theta$) estimated using randomly picked $h-1$ data points. In the remainder of the text we adopt this tensor theoretic approach. 

The sampling strategy used to construct the sample matrix $H$ critically affects the clustering and thus, overall performance of the model fitting solution.  

\subsection{Why distribution of sampling is important?}
In tensor theoretic approach, pairwise affinity matrix $G$ is constructed by multiplying the matrix $H$ with its transpose where $H(i,l) = e^{-r^2_{\theta_l}(i) /2\sigma^2}$, $r^2_{\theta_l}(i)$ is the squared residual of point $i$ to model $\theta_l$ (obtained by fitting to a tuple $\tau_l$) and $\sigma$ is a normalization constant. 
\begin{equation}
	G_{[N \times N]} = HH^\top = \sum_{l=1}^{n_H} \underbrace{\left[H^{(l)}{H^{(l)}}^\top \right]}_{G^{(l)}_{[N \times N]}}
	\label{equ:graphCont}
\end{equation}
where  $H^{(l)}$ is the $l^{th}$ column of $H$ corresponding to the hypothesis  $\theta_l$, $G^{(l)}$ is the contribution of hypothesis  $\theta_l$ to the overall affinity matrix ($G$) and $n_H$ is the total number of hypotheses.

\begin{figure}
	\centering
	\subfloat[Data] {
		\includegraphics[width=3.25in]{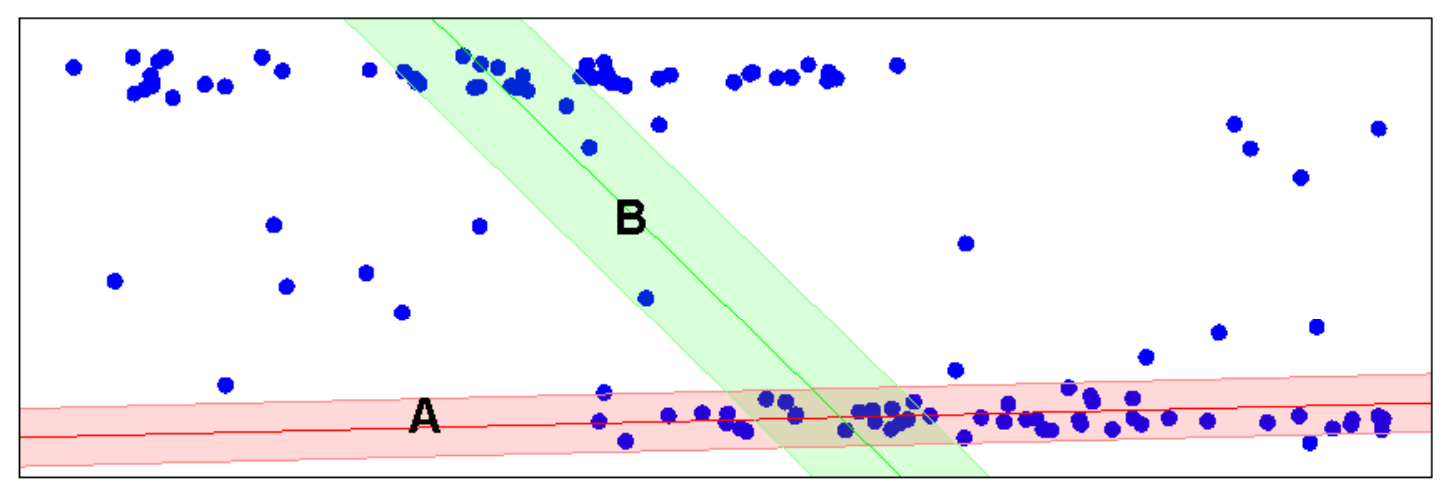}
		\label{fig:lineEx:subfig1}
	}
	
	\subfloat[$G^{(A)}$]{
		\includegraphics[width=1.6in]{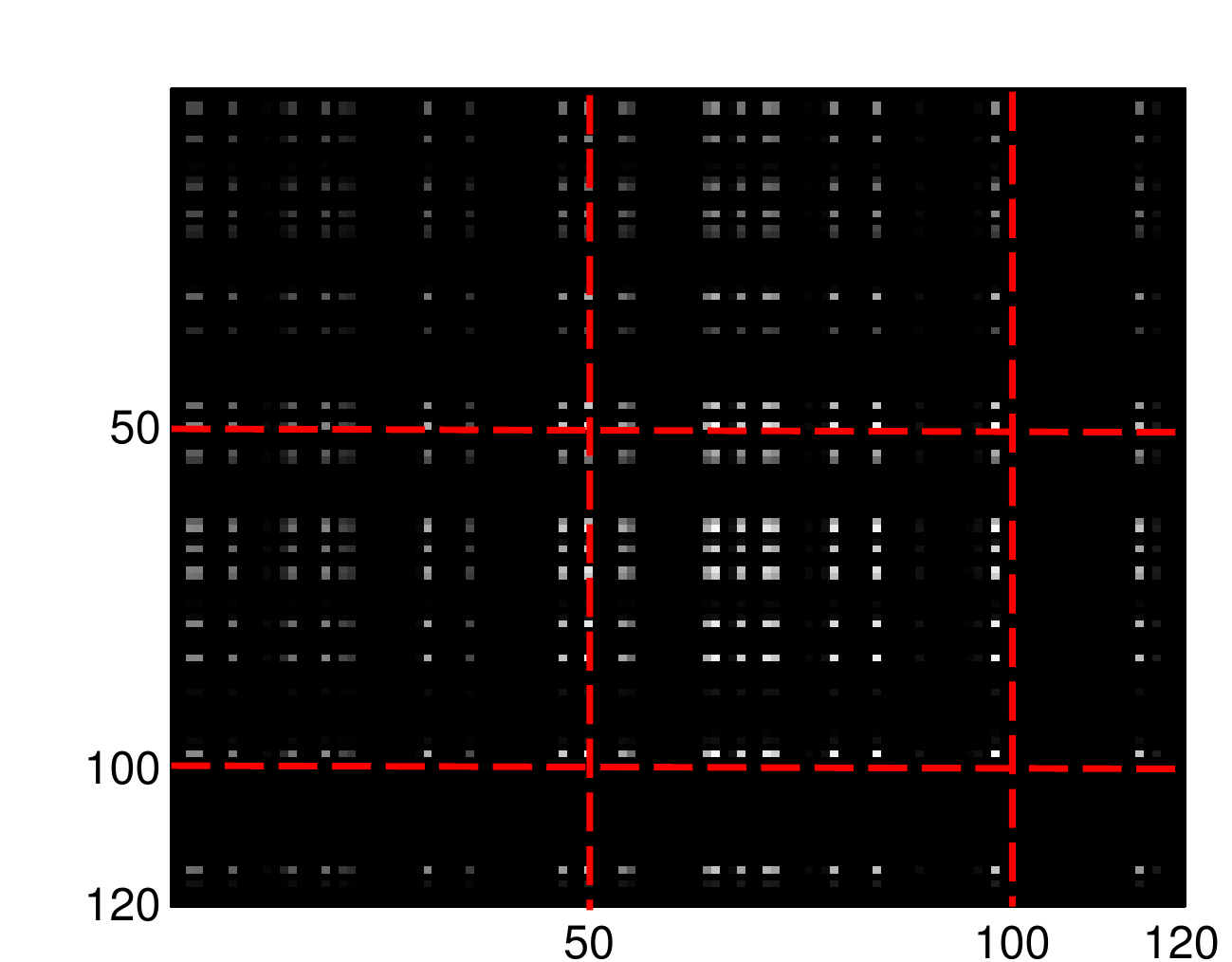}
		\label{fig:lineEx:subfig2}
	}
	\subfloat[$G^{(B)}$]{
		\includegraphics[width=1.6in]{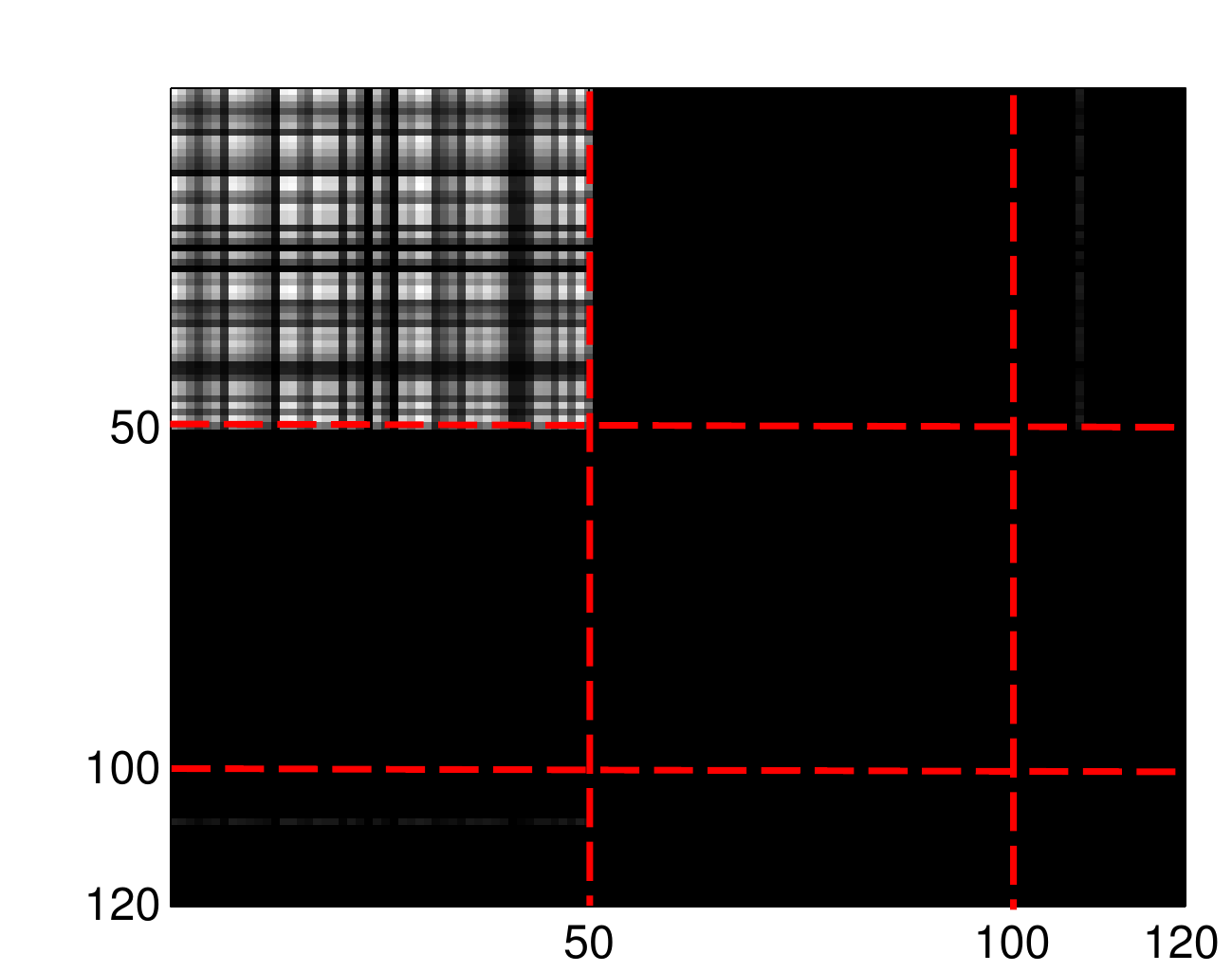}
		\label{fig:lineEx:subfig3}
	}

	\caption{An example line fitting scenario on a synthetic dataset containing two lines and some outliers. The lines A and B show two model hypotheses while the shaded areas around the lines indicate to the corresponding $\sigma$ values. (b) and (c) show the contributions of hypotheses A and B to the overall graph respectively. The data points are sorted according to their model affiliation, where the first 50 data points belong to line one followed by line two (50 points) and the outliers (20 points). The dashed lines indicate the cluster boundaries.}
	\label{fig:lineEx}
\end{figure}

When a model hypothesis $\theta_l$ is close to an underlying structure in data (Hypothesis A in \figref{lineEx:subfig1}), the inlier points of that structure would have relatively small residuals and the resulting $G^{(l)}$ (\figref{lineEx:subfig2}) would have high affinities between the inliers and low affinity values for all other point pairs (outlier-outlier, outlier-inlier). On the other hand, when a model hypothesis $\theta_l$ is far (in parameter space) from any underlying structure, the presumption is that the resulting residual would be large, leading to a $G^{(l)} \approx \textbf{0}_{[N \times N]}$. However, as seen in \figref{lineEx:subfig1} (for Hypothesis B), this is not always the case in model fitting. It is highly likely that there exists some data points that give small residuals even for such hypothesis (far from any underlying model) leading to high $H(i,l)$ values. The resulting $G^{(l)}$ (\figref{lineEx:subfig3}) would have high affinities between some unrelated points that can be seen as noise in the overall graph. The effect of these bad hypothesis can be amplified by the fact that the normalization factor, $\sigma$ is often overestimated (using robust statistical methods) when the hypothesis $\theta_l$ is far (in parameter space) from any underlying structure. It is important to note that if none of the hypotheses (used in constructing the graph) are close to a underlying structure, then the overall graph would not have higher affinities between the data points in that structure and the clustering methods would not be able to segment that structure. 

The above example shows that the sampling process influences the level of noise in the graph. While spectral clustering can tolerate some level of noise, it has been proved that this noise level is related to the size of the smallest cluster we want to recover (tolerable noise level goes up rapidly with the size of the smallest cluster) \cite{Balakrishnan2011}. As model fitting often involves recovering small structures, it is highly important to limit the noise level in the affinity matrix.

For any two data points $x_i,x_j$ we can write:
\begin{equation}
	G(i,j) = \frac{1}{n_H} \sum_{l=1}^{n_H} \underbrace{e^{-\frac{\left (r^2_{\theta_l}(i) + r^2_{\theta_l}(j)\right )}{2\sigma^2}}}_{g_{ij}(\theta_l)} \xrightarrow[n_H\uparrow]{as} \int P_\theta \cdot g_{ij}(\theta_l)~d\theta
	\label{equ:gij}
\end{equation}
For any model fitting problem with $p > 2$ there exists infinite number of models $\theta_l$ where  $g_{ij}(\theta_l) \rightarrow 1$. This implies that for any two points, $G(i,j)$ (according to \eqnref{gij}) can be maximized or minimized by choosing $P_\theta$ accordingly.


For a graph to have the block diagonal structure suitable for clustering, $G(i,j)$ needs to be large for $x_i \wedge x_j \in \theta_t$ and small otherwise. If hypotheses are selected from a Gaussian mixture distribution with sharp peaks around the underlying model parameters and low density in other places and $\theta_t$ representing the true underlying structures, we have:
\begin{equation}
	P_\theta = \sum_{t=1}^{m} \phi_t ~\mathcal{N} (\theta_t, \Sigma_t ).
\end{equation}
the edge weights approach the following values when $\Sigma_t \to \textbf{0}$:
\begin{equation}
	G(i,j) \to \left\{\begin{matrix}
		\phi_t & i \wedge j \in \theta_t\\ 
		0 & i \wedge j \notin \theta_t
	\end{matrix}\right.
\end{equation}
The $G$ results in a graph that has a block diagonal structure suitable for clustering. Of course, generating sample hypotheses form this distribution is not possible because it is unknown until the problem is solved. 

\begin{figure*}[!t]
	\centering
	\subfloat[Data]{
		\includegraphics[width=1.5in] {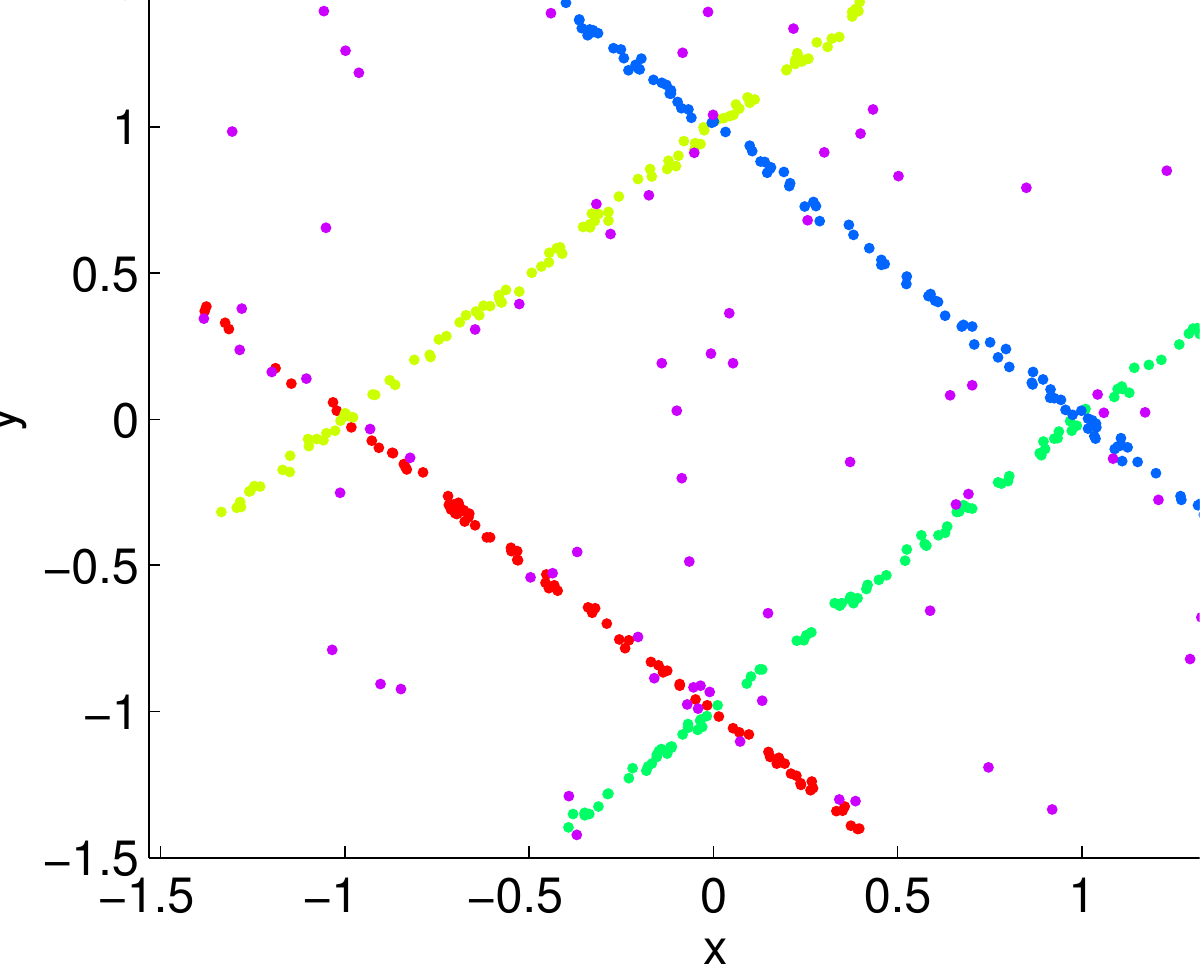}
		\label{fig:EstimationErrorCDF:subfig1}
	}
	\subfloat[Cost-based sampling]{
		\includegraphics[width=1.5in] {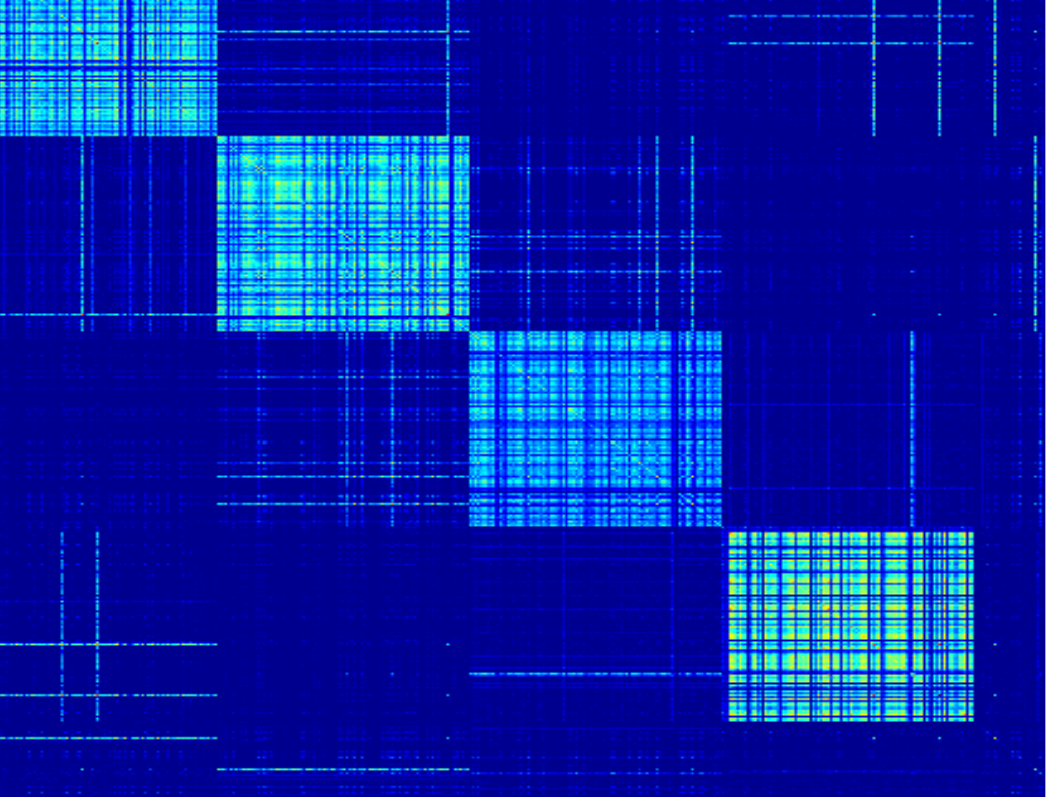}
		\label{fig:EstimationErrorCDF:subfig2}
	}
	\subfloat[Uniform sampling]{
		\includegraphics[width=1.5in] {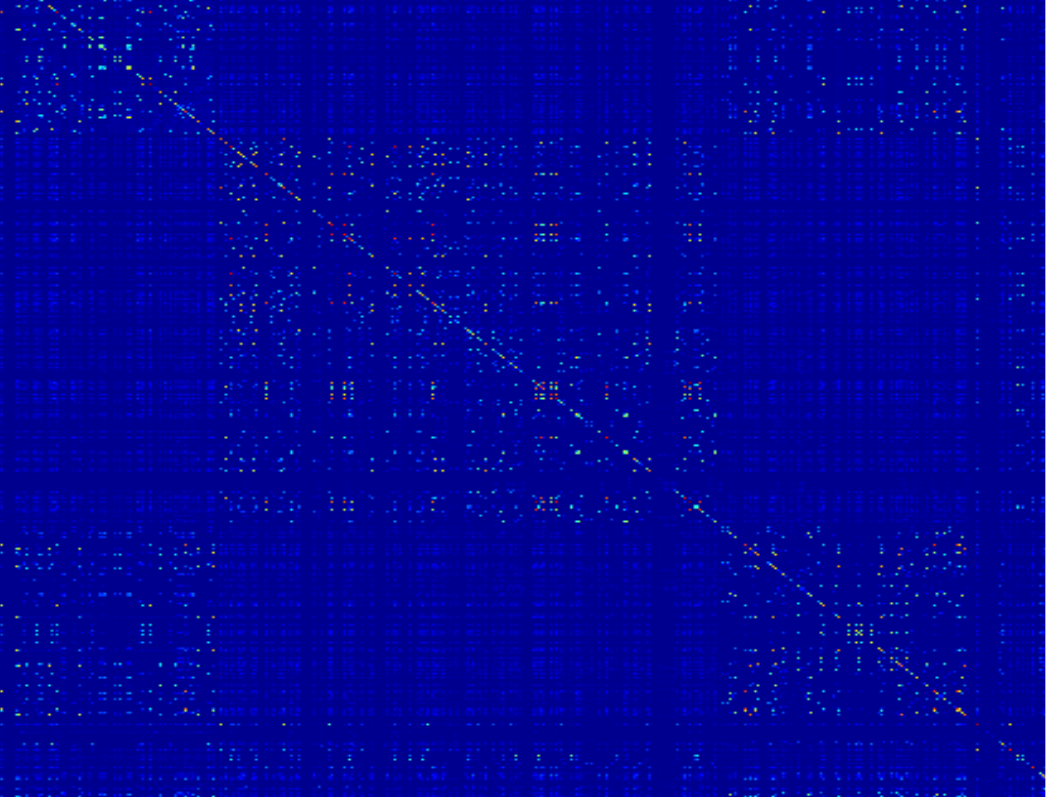}
		\label{fig:EstimationErrorCDF:subfig3}
	}
	\subfloat[Random sampling]{
		\includegraphics[width=1.5in] {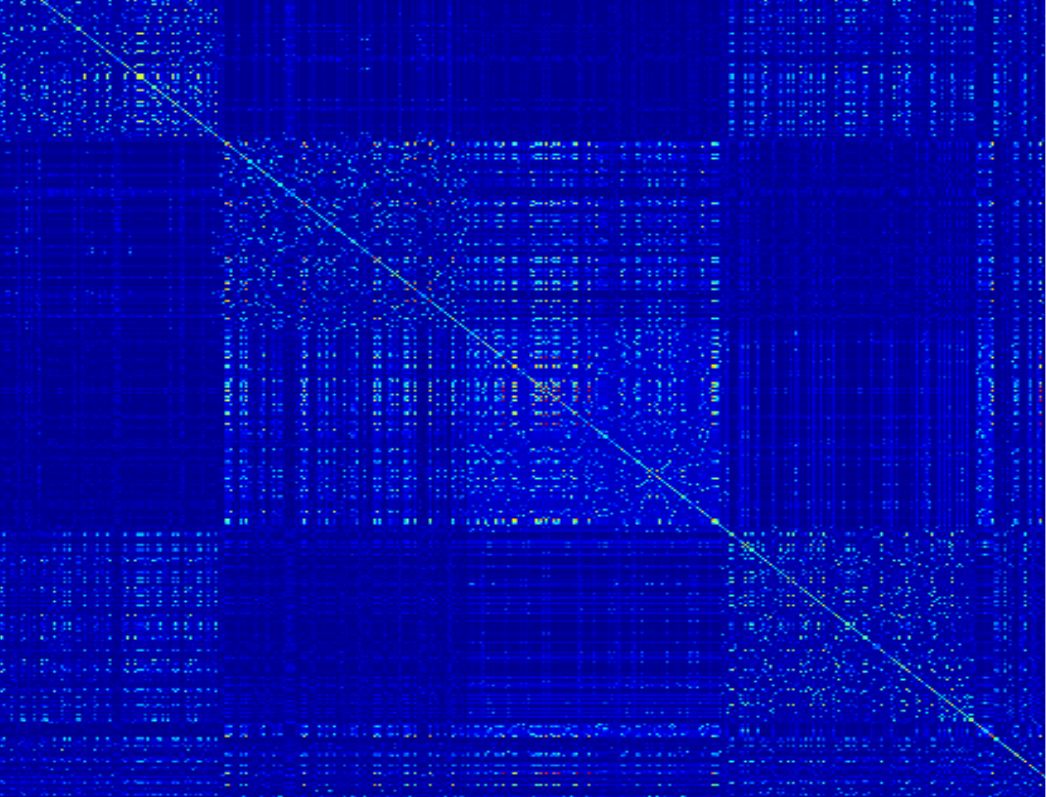}
		\label{fig:EstimationErrorCDF:subfig4}
	}

	\subfloat[Cost-based sampling]{
		\includegraphics[width=1.5in] {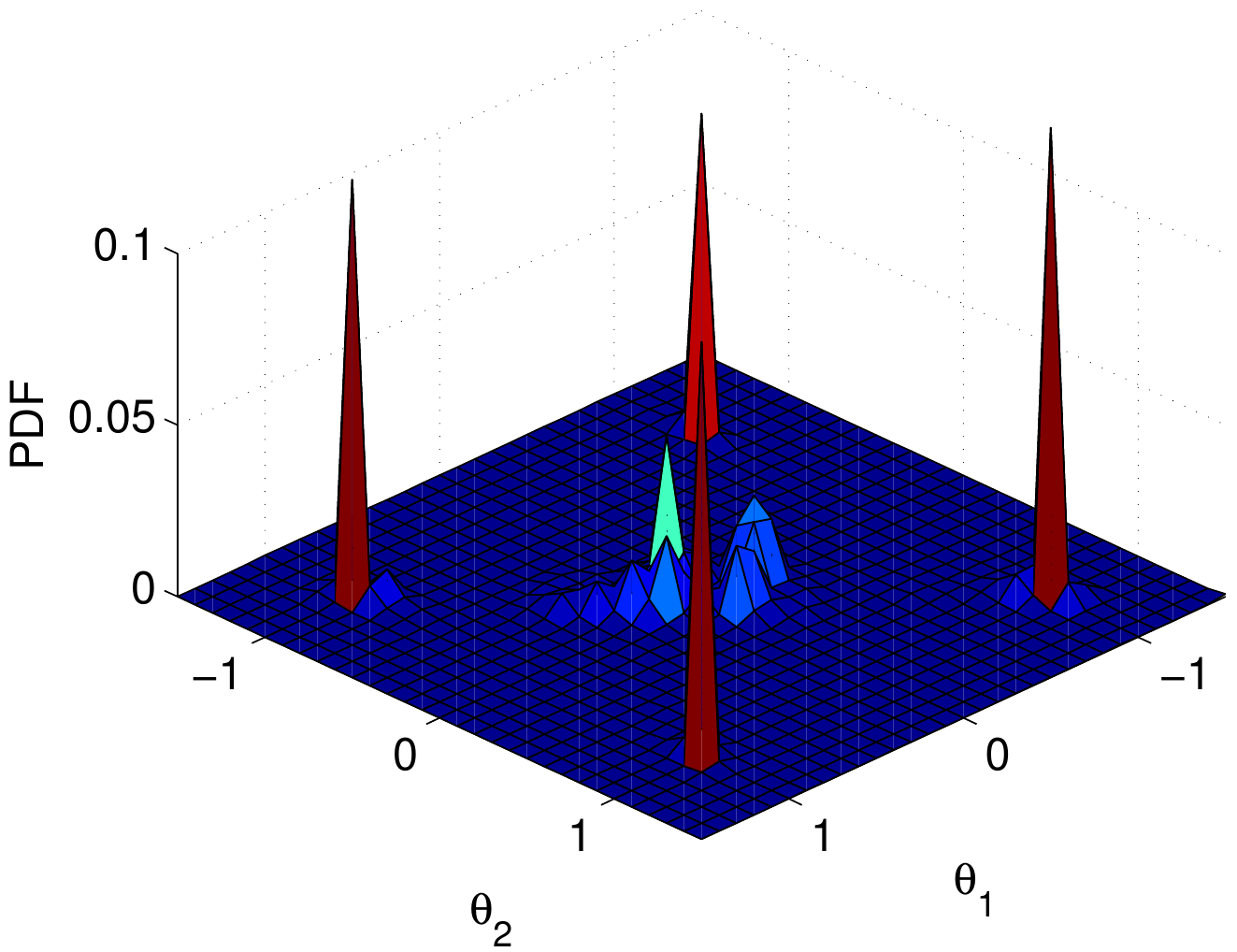}
		\label{fig:EstimationErrorCDF:subfig5}
	}
	\subfloat[Uniform sampling]{
		\includegraphics[width=1.5in] {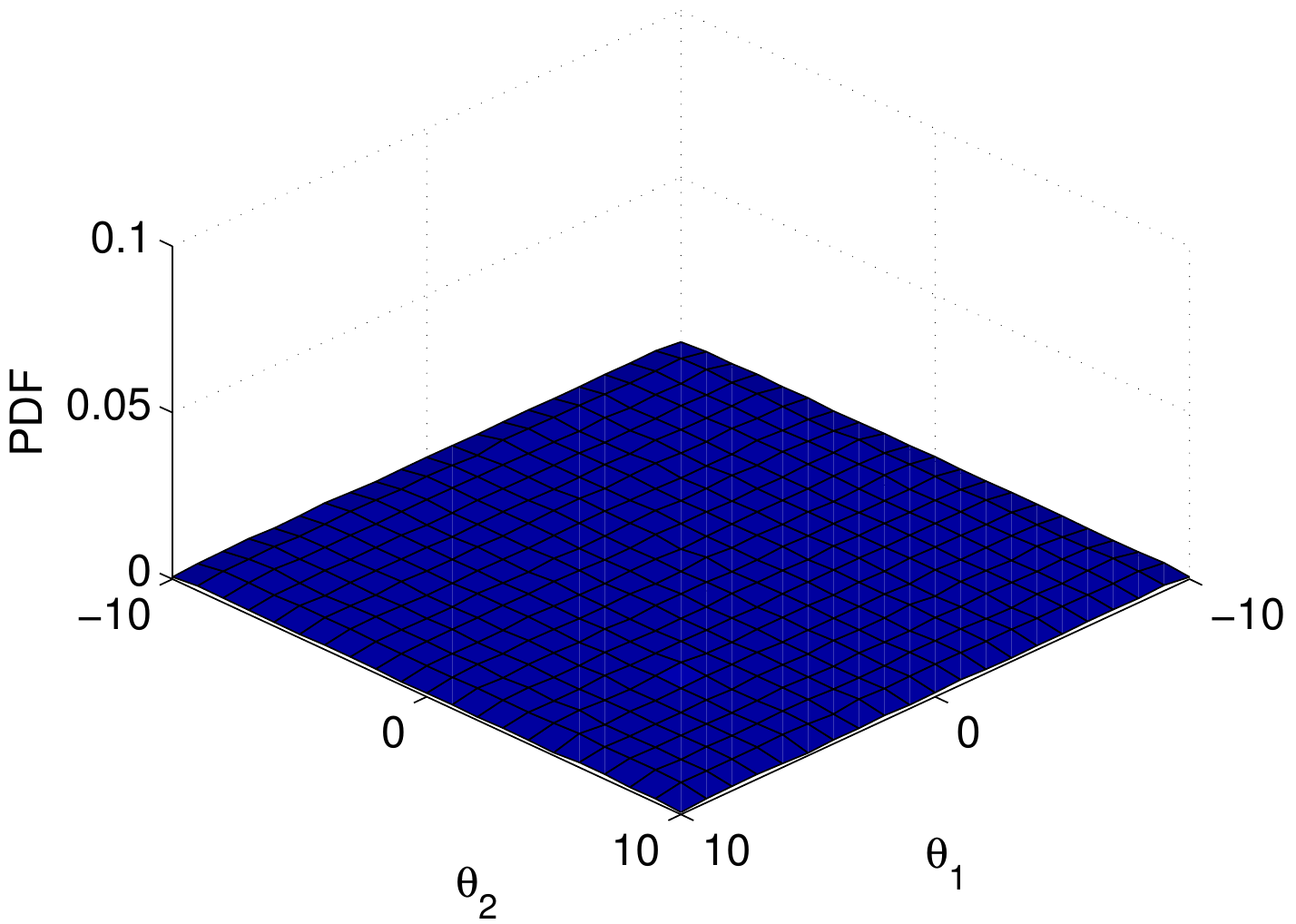}
		\label{fig:EstimationErrorCDF:subfig6}
	}
	\subfloat[Random sampling]{
		\includegraphics[width=1.5in] {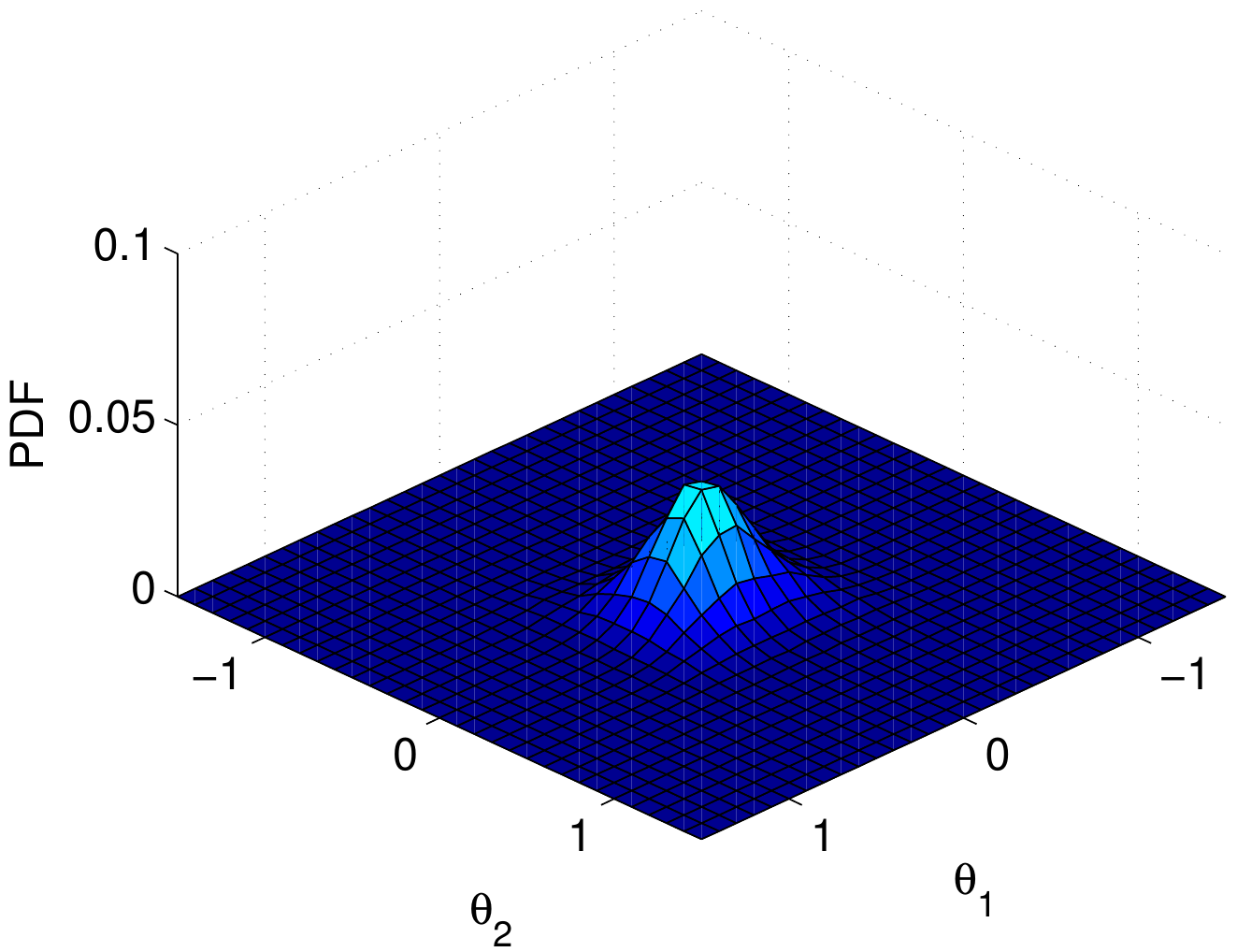}
		\label{fig:EstimationErrorCDF:subfig7}
	}	
	
	\caption{The synthetic dataset containing four line structures is shown in (a) while the graphs produced by the  cost based sampling, random and uniform sampling (-10,10) methods are shown in (b-d) respectively. The respective hypothesis distributions are shown in (e-f). While the CBS method has resulted in a graph favorable for clustering the other two sampling strategies have produced graphs with little information.}
	\label{fig:examleGraph}
\end{figure*}

This point is further illustrated using a simple model fitting experiment using a synthetic dataset containing four lines. Each line contain 100 data points with additive Gaussian noise $\mathcal{N}(0, 0.02^2)$, while 50 gross outliers were also added to those lines. First, $500$ hypotheses were generated using uniform sampling, random sampling (using 5-tuples) and the sampling scheme proposed in this paper (CBS). These hypotheses were then used to generate the three graphs shown in \figref{examleGraph}. As the data is arranged based on the structures membership, a properly constructed graph should show a block diagonal structure with high similarities between points in the same structure and low similarities for data from different structures.The figure shows that while the CBS method has resulted in a graph favorable for clustering the other two sampling strategies have produced graphs with little information. The corresponding hypothesis distributions (\figref{examleGraph} (e-f)) show that only CBS has generated high amount of hypotheses closer to the underlying structure.

Govindu \cite{Govindu2005} used randomly sampled $h-1$ (for affinities of order $h$) data points and calculated a column of $H$ by computing the affinity from those to each point in the dataset. It is well known that the probability of obtaining a clean sample, leading to a hypothesis close to a true structure in data, decreases exponentially with the size of the tuple \cite{Agarwal2005}. Hence it becomes increasingly unlikely to obtian a good graph for models with high number of parameters using random sampling.   

There are several techniques in the literature that try to tackle the clustering problem by tapping into available information regarding the likelihood distribution of good hypotheses. For instance, spectral curvature clustering \cite{Chen2009}, which is an algorithm designed for affine subspace clustering, employs an iterative sampling mechanism that increases the chance of finding good hypotheses. In this scheme, a randomly chosen affinity matrix ($H$) is used to build a graph and partitions it using the spectral clustering method to generate an initial segmentation of the dataset. Data points within each segment of this clustering are then sampled to generate a new set of columns of $H$. This process is repeated several times to improve the final clustering results.          

Similarly, Ochs and Brox \cite{Ochs2012} used higher order affinities in a hyper-graph setting for motion segmentation of video sequences. In their method, the affinity matrix is obtained using a sampling strategy that is partly random and partly deterministic. The higher order affinities are based on 3-tuples generated by choosing two points randomly. The third points are then chosen as a mixture of $12$ nearest neighbor points and $30$ random 3rd points.  

The previous guided sampling approaches generate the columns of the affinity matrix using the minimal size tuples. Purkait \etal \cite{Purkait2014} advocated the use of larger tuples and showed that if those tuples are selected correctly, the hypotheses distribution would be closer to the true model parameters compared to smaller tuples. However selecting larger all inlier (correct) tuples using random sampling is highly unlikely. Purkait \etal \cite{Purkait2014} suggested to use Random Cluster Models (RCM) \cite{Swendsen1987} to improve the sampling efficiency. RCM is based on selecting the tuples iteratively in a way that at every iteration the samples are selected using the segmentation results obtained by enforcing the spatial smoothness on the results of the previous iteration. This approach is particularly advantageous where the application satisfies the spatial smoothness requirements. Our proposed approach for constructing the affinity matrix, without relying on the existence of spatial smoothness, is explained in the next section.

\section{Proposed Method}
\label{sec:method}
This section describes a new approach for multi-structural model fitting problem. 
Similar to \cite{Agarwal2005}, \cite{Govindu2005}, we approach multi-structural fitting as a clustering problem with the intention of applying spectral clustering. In this approach, the pairwise affinity matrix $G$ for spectral clustering is obtained by projecting the higher order affinity tensor ($\mathcal{P}$) via multiplying an approximated flattened matrix $H$ with its transpose. For affinities of order $h$, each column of $H$ is obtained by sampling $h-1$ data points and calculating the affinity of each point to those sampled points. The affinity of a data point $i$ to a $h-1$ tuple is calculated as $e^{-r^2_{\theta_l} (i)/(2\sigma^2)}$ where $\theta_l$ is the model parameters fitted to $h-1$ tuple and $\sigma$ is the normalization factor. For the sake of clarity, in the remainder of this text, a $h-1$ tuple ($\tau_l$) used to generate a column of $H$ is referred to as an edge while its respective model ($\theta_l$) is called a hypothesis. 

As discussed in \secref{background}, the way we sample the edges affects the information content of the resulting graph and our ultimate goal is to sample edges in such a way that the distribution of their associated hypotheses resembles the true distribution of the model parameters. While the true distribution of the model parameters for a given dataset $p(\theta~|~X)$ is unknown until the problem is solved, using Bayes' theorem it can be written as follows:

\begin{equation}
	p(\theta|X) \propto p(X|\theta) p(\theta)
\end{equation}
where $p(X|\theta)$ is the likelihood of observing data $X$ under the model $\theta$ and $p(\theta)$ is the prior distribution of $\theta$. Given that the prior is uninformative (i.e. any parameter vector is equally likely), the posterior is largely determined by the data (the posterior is data-driven) and can be approximated by: $p(\theta|X) \propto  p(X|\theta)$.

A robust objective function is often used in multi-structural model fitting applications to quantify the likelihood of existence of a structure in data \cite{Stewart1997}. On that basis, we would argue that it can be a good approximation of the model parameters likelihood. For example the sample consensus objective function as employed in RANSAC is expected to have a peak in places where a true structure is present (in the parameter space) and low values where there are no structures. It should be noted here that when there are structures of different size, the sample consensus function associates higher values for larger structures (hence it is biased towards large structures). In this work, we select the cost function of the least k-th order statistics (LkOS) estimator as the objective function, as it has shown to perform with stability and high breakdown point \cite{Rousseeuw2005} in various applications and it is not biased towards large structures (LkOS is biased towards structures with low variance, which is a desirable property). A modified version of the LkOS cost function used in \cite{Bab-Hadiashar2008} is as follows:
\begin{equation}
	C({\theta}) =  \sum_{j=0}^{p-1}r_{i_{j-m,\theta}}^2(\theta)
	\label{equ:kthSortedCostFunction}
\end{equation}
where $r_{i}^2(\theta)$ is the $i$-th sorted squared residual with respect to model $\theta$ and $i_{k, \theta}$ is the index of the $k$-th sorted squared residual with respect to model $\theta$. Here $k$ refers to the minimum acceptable size of a structure in a given application and its value should be significantly larger than the dimension of the parameter space ($k \gg p$). Because the above cost function is designed to have minima around the underlying structures, the model parameters likelihood function can be expressed as:
\begin{equation}
	P_\theta \propto p(X|\theta)  \approx \frac{1}{Z} e^{-C({\theta})} .
	\label{equ:SamplingDist}
\end{equation}
The above function is highly non-linear and its evaluation over the entire parameter space, required for calculating the normalizing constant $Z$, would not be feasible. The common approach for sampling from a distribution that can only be evaluated up to a proportional constant on specified points is to use the Markov Chain Monte Carlo (MCMC) method (e.g. by using Metropolis-Hasting algorithm). However such algorithms need a good update distribution to be effective, and simple update distributions like random walk would be inefficient and may not traverse the full parameter space \cite{Andrieu2003}. In particular, setting up random walk distributions need the information regarding the span of model parameters, which is unknown until the problem is solved.

\subsection{Sampling edges using the robust cost function }
Using derivatives of the order statistics function in \eqnoref{kthSortedCostFunction}, a greedy iterative sampling strategy was proposed in \cite{Bab-Hadiashar2008} that is intentionally biased towards generating data samples from a structure in the data. This sampling strategy was then used to generate putative model hypotheses for different size tuples in conjunction with the fit and remove strategy to recover multiple structures in data \cite{Tennakoon2015}, \cite{Bab-Hadiashar2008}. Because fit and remove strategy is susceptible to errors in the initial stages, the sampling had to be reinitialized (randomly) several times to reduce the probability of error propagation in the sequential fit and remove stages. 

In this paper, we propose a modified version of this iterative update procedure (recalled in Algorithm~\ref{alg:HMSS}) to generate model estimates (edges) that are close to the peaks of the true parameter density function $p(\theta|X)$. Each edge used in constructing the $H$ matrix of the proposed method is obtained as follows:  Initially a $h$-tuple ($h = p+2$) is picked according to the inclusion weights $W$ (this will be explained later). Using this tuple as the starting point the following update is run until convergence. A model hypothesis is generated using the selected tuple, and the residuals form each data point to this hypothesis are calculated. These residuals are then sorted and the $h$ points around the $k$-th sorted index are selected as the updated tuple for next iteration. 

In practice, the above update step has the following property: If the current $h$-tuple is a clean sample (all inliers) from a structure in data, there is a high probability that the next sample will also be from the same structure as there should be at least $k$ points agreeing to each true structure. On the other hand if the current hypothesis is not supported by $k$ points (not a structure in data), the next hypothesis would be at a distance in the parameter space. It is shown that residuals of a data structure with respect to an arbitrary hypothesis have a high probability of clustering together in the sorted residual space \cite{Toldo2008}, \cite{Haifeng2003}. As the next sample is selected from the sorted residual space, the probability of hitting a clean sample would then be higher than selecting it randomly.  

Following \cite{Tennakoon2015}, we use the following criterion to decide whether the update procedure is converged to a structure in data: 
\begin{equation}
	\begin{split}
		F_{stop} = \left (  r_{i_{k,\theta_l}}^2(\theta_l) < \frac{1}{h}\sum_{j=k-h+1}^{k}\underbrace{{r_{i_{j,\theta_{(l-1)}}}^2(\theta_l)}}_\text{(a)}  \right )  \wedge \\
		\left (  r_{i_{k,\theta_l}}^2(\theta_l) < \frac{1}{h}\sum_{j=k-h+1}^{k}\underbrace{{r_{i_{j,\theta_{(l-2)}}}^2(\theta_l)}}_\text{(b)} \right ).
	\end{split}
	\label{equ:stopCriterion}
\end{equation} 
Here $(a)$ and $(b)$ are the squared residuals of the edge points in iterations $l-1$ and $l-2$ with respect to the current parameters $\theta_l$. This criterion checks the data points associated with the two previous samples to see if the average residuals of those points (with respect to the current parameters) are still lower than the inclusion threshold associated with having $k$ points (assuming that a structure has at least $k$ points implies that data points with residuals less than $ r_{i_{k,\theta_l}}^2(\theta_l)$ are all inliers). This indicates that the samples selected in the last three iterations are likely to be from the same structure hence the algorithm has converged.

\renewcommand{\algorithmicrequire}{\textbf{Inputs:}}
\renewcommand{\algorithmicensure}{\textbf{Output:}}
\newcommand{\LINEIF}[2]{%
	\STATE\algorithmicif\ {#1}\ \algorithmicthen\ {#2} \algorithmicend\ \algorithmicif%
}

\begin{algorithm}[!t]                      
	\caption{Step-by-step algorithm of sample generation ($runCBS\_SG$)}          
	\label{alg:HMSS}                           
	\begin{algorithmic} [1]
		\REQUIRE Data Points (${X} \in  \left [ x_i \right ]_{i=1}^N$), minimum cluster size ($k$), $T$, inclusion weights ($W$) 
		\ENSURE Final data indexes $I_{l}$, Scale $\sigma$ 
		\STATE $l_{max} \gets 50$, $h \gets p+2$, $l \gets 0$
		
		\STATE Select a $h$-tuple ($I_{0}$) from the data points according to weights $W$. 
		\STATE Generate model hypothesis $\theta_0$ using the $h$-tuple $I_{0}$.
		\REPEAT 
		\STATE $[r^2(\theta_l), i_{\theta_l}] = $SortedRes(${X}, \theta_l)$.
		\STATE $I_{l+1} \gets [x_{i_{\theta_l}(j)}]_{j=k-h+1}^{k}$
		\STATE $\theta_{l+1} \gets$ LeastSquareFit$\left ( I_{l+1} \right )$
		\STATE Evaluate the stopping criterion ($F_{stop}$) 
		\LINEIF{$F_{stop}$}{break}
		\UNTIL{$( l{++} > l_{max}) $}
		\STATE $\sigma \gets$ MSSE($X,\theta_{l}, k , T$)
	\end{algorithmic}
\end{algorithm}

\subsection{Sub-sampling data}
Although the above update procedure has a high probability of generating an edge that results in a hypothesis close to a peak in $p(\theta|X)$, there is no guarantee that all the structures present in the data will be visited given that the update step is reinitialized from random locations. If some of the structures were not visited by the sampling procedure, the resulting graph would not contain the information required to identify those structures.

To ensure that the algorithm would visit all the structures in data, we propose to use a data sub-sampling strategy. Each run of the the update procedure in Algorithm~\ref{alg:HMSS} is executed only on a subset of data selected based on an inclusion weight ($W$). The inclusion weight, which is initialized to one, is designed in such a way that at every iteration, it will give higher importance to data points that are not modeled by the hypothesis used in the previous iterations. This will progressively increase the chance of unmodeled data to be included in the sampling process. This idea is similar to the Bagging predictors \cite{Breiman1996} with boosting \cite{Freund1996},\cite{Freund1997} in machine learning. In Bagging predictors multiple subsets of data formed by bootstrap replicates of the dataset are used to estimate the models, which are then aggregated to get the final model. Boosting improves the bagging process by giving importance to unclassified data points in successive classifiers. 

The overall edge generation procedure is as follows: A data subset of size $N_s$ is sampled from data using the inclusion weights $W$ without replacement ($W$ is normalized in $sampleData(\cdot)$ function). This sub-sample is then used in the update procedure in algorithm~\ref{alg:HMSS}, which produces an edge. Next the inclusion weights $W$ of the inliers to the above hypothesis are decreased while the inclusion weights of the remaining points are increased. This process is repeated for a fixed number of iterations. 
The complete steps of the proposed method (CBS) are listed in Algorithm~\ref{alg:PM}. 

The scale of noise plays a crucial role in the success of segmentation methods. In spectral clustering based model fitting methods, the scale is used to convert the residuals to an affinity measure. While most competing algorithms require this as an input parameter \cite{Purkait2014}, \cite{Pham2014}, the proposed method estimates the scale of noise from the given data. In this implementation, we selected the MSSE \cite{Bab-Hadiashar1999} to estimate the scale of noise. The MSSE algorithm requires a constant threshold $T$ as an input. This threshold defines the inclusion percentage of inliers. Assuming a normal distribution for noise, it is usually set to 2.5, i.e. $T=2.5$ will include 99\% of normally distributed inliers. Desirable properties of this estimator for dealing with small structures were discussed in \cite{Hoseinnezhad2010}.      

\begin{algorithm}[!t]                      
	\caption{Step-by-step algorithm of proposed model-fitting methods}          
	\label{alg:PM}                           
	\begin{algorithmic} [1]
		\REQUIRE Data Points ($X_{d \times N} \gets  \left [ x_i \right ]_{i=1}^N$), minimum size of structure ($k$), Number of structures ($n_c$), number of hypothesis ($n_H$), $T \gets [2.0 \sim 3.5]$
		
		\STATE $W \gets [\frac{1}{N} \dots \frac{1}{N}]_{1 \times N}$; $N_s \gets N/n_c$; $w \gets \frac{20}{N}$
		\REPEAT
		\STATE Sample $N_s$ data points from $X$ based on inclusion weights $W$; $[X_s, W_s] \gets sampleData(X, W, S_f)$.
		
		\STATE $[I_s, \sigma] \gets runCBS\_SG(X_s, k, T, W_s)$
		\STATE Calculate residuals ($r_{I_s}^2$) to all data points from the h-tuple $I_s$. 
		\STATE $H(:,i) \gets exp(-r_{I_s}^2/2\sigma_i^2)$
		\STATE Calculate inliers $C_{inl}$ using $r_{I_s},\sigma_i$.
		\STATE $W \gets W \times 2$
		\STATE $W(C_{inl}) \gets W(C_{inl}) \div 4$
		\STATE $W(W>w) \gets {1}/{N}$
		\STATE $W \gets {W}/{sum(W)}$
		\UNTIL{$i{++} > n_H$}
		\STATE $G \gets H  H^\top$
		\STATE $[labels] \gets spectralClustering(G, n_c)$
	\end{algorithmic}
\end{algorithm}

\section{Experimental Results}
\label{sec:experimentalanalysis}
We have evaluated the performance of the proposed method for multi-object motion segmentation in several well-known datasets. The results of the proposed cost-based sampling method (CBS) were then compared with state-of-the-art robust multi-model fitting methods. The selected methods use higher order affinities Spectral Curvature Clustering (SCC \cite{Chen2009}, HOSC\cite{Purkait2014} and OB \cite{Ochs2012}) or are based on energy minimization (RCMSA \cite{Pham2014}, PEARL \cite{Boykov2001} and QP-MF \cite{Jin2011}).  

The accuracy of all methods was evaluated using the commonly used clustering error  (CE) measure given in \cite{Purkait2014}:
\begin{equation}
	CE = \min_{\Gamma} \frac{\sum_{i=1}^{N} \delta \left( L^*(i) \neq L_r^\Gamma (i) \right) }{N} \times 100
	\label{equ:custeringError}
\end{equation}
where $L^*(i)$ is the true label of point $i$, $L_r(i)$ is the label obtained via the method under evaluation and $\Gamma$ is a permutation of labels. The function $\delta(\cdot)$ returns one when the input condition is true and zero otherwise.

The proposed CBS algorithm was coded in MATLAB (The code is publicly available: https://github.com/RuwanT/model-fitting-cbs) and the results for competing methods were generated using the code provided by the authors of those works. 
The experiments were run on a Dell Precision M3800 laptop with Intel i7-4712HQ processor.

\subsection{Analysis of the proposed method}
In this section we investigate the significance of each part of the proposed algorithm and the effect of its parameters on its accuracy. This analysis was conducted using a Two-view motion segmentation problem (see \secref{twoviewMS} for more details).  

\begin{figure*}[!t]
	\centering
	\subfloat[Ground truth]{
		\includegraphics[height=1.25in,width=1.5in] {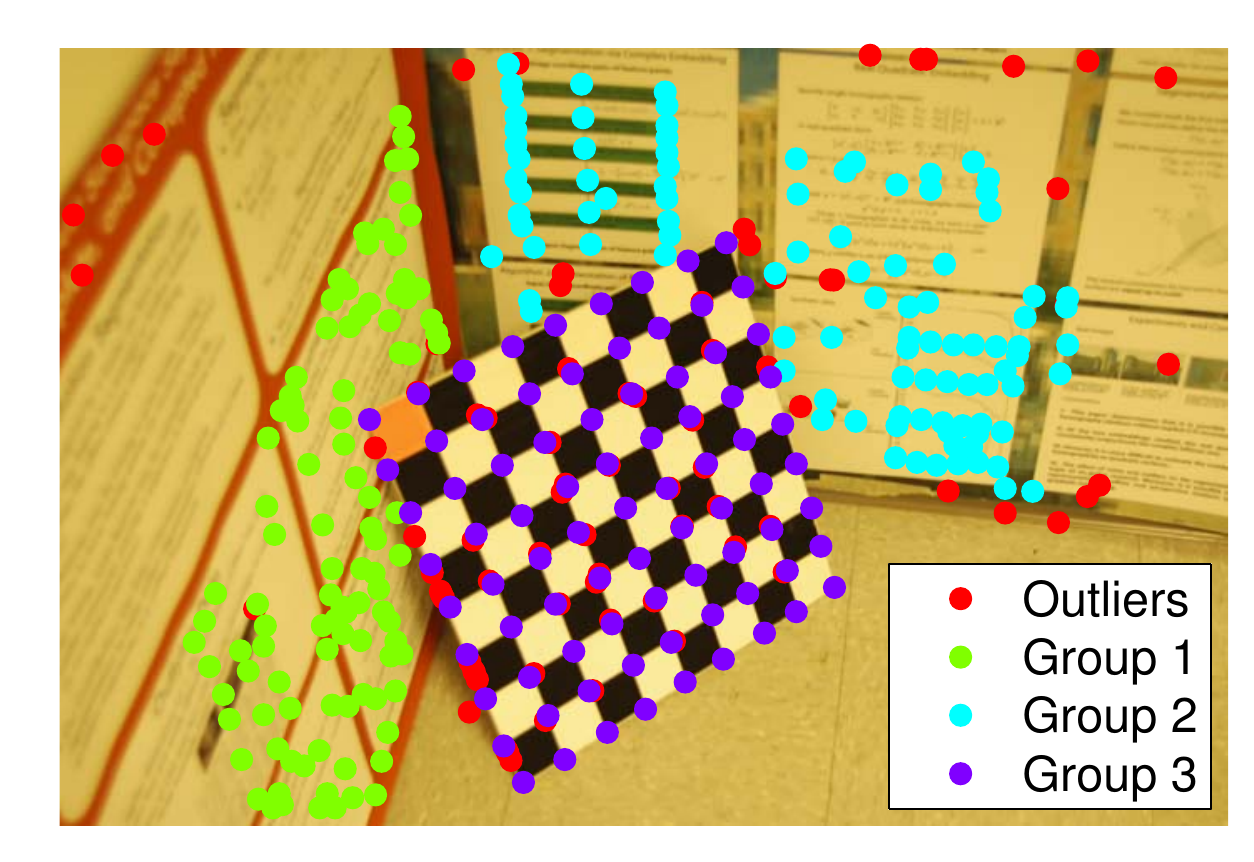}
		\label{fig:posterCheckerboard:subfig1}
	}
	\subfloat[Random]{
		\includegraphics[height=1.25in,width=1.5in] {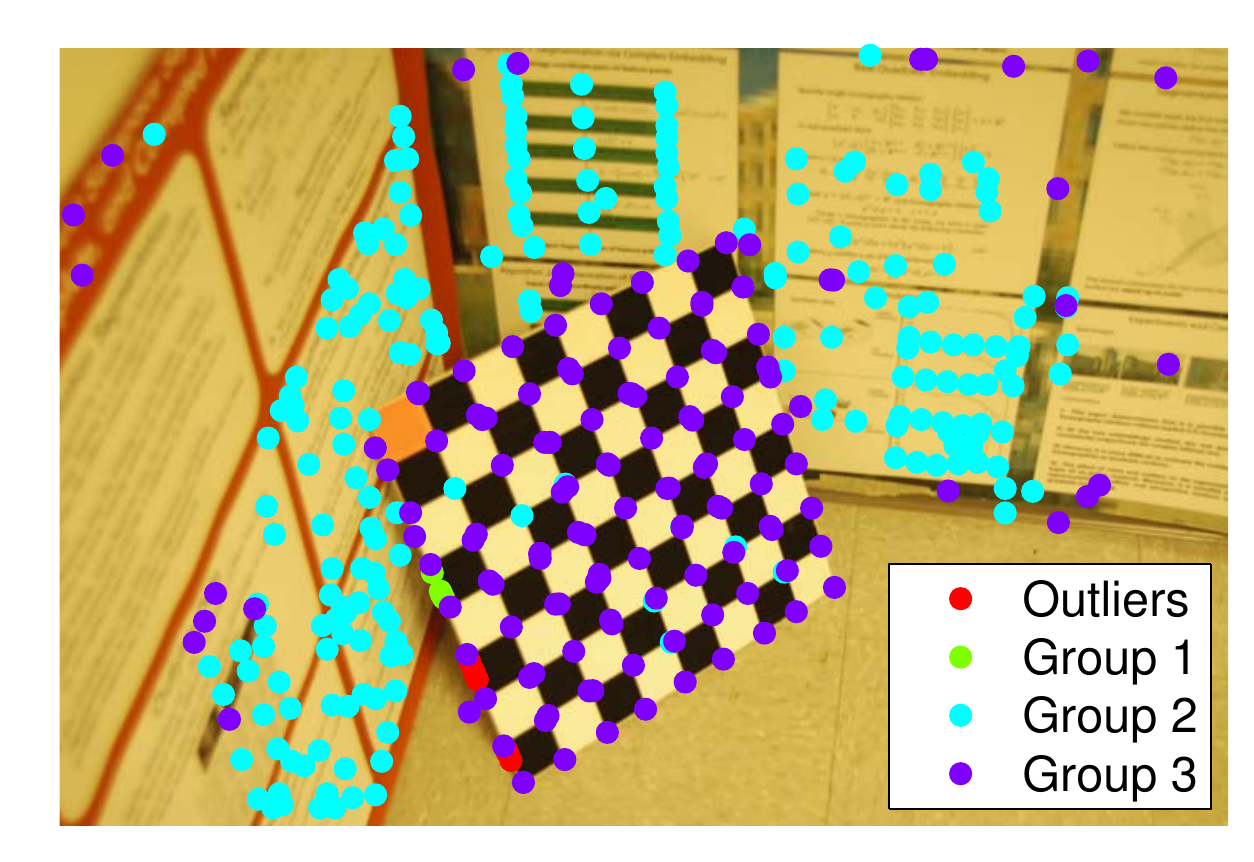}
		\label{fig:posterCheckerboard:subfig2}
	}
	\subfloat[CBS-nSS]{
		\includegraphics[height=1.25in,width=1.5in] {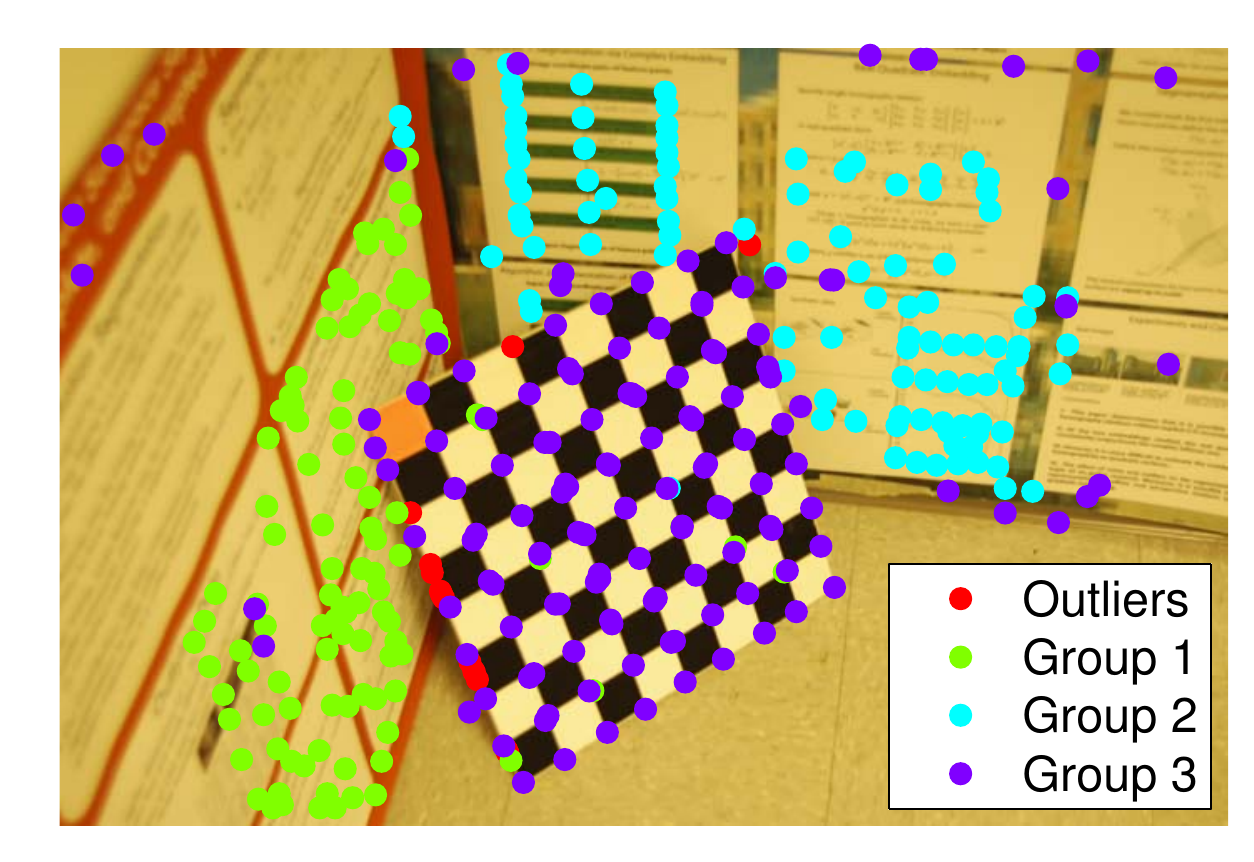}
		\label{fig:posterCheckerboard:subfig3}
	}
	\subfloat[CBS]{
		\includegraphics[height=1.25in,width=1.5in] {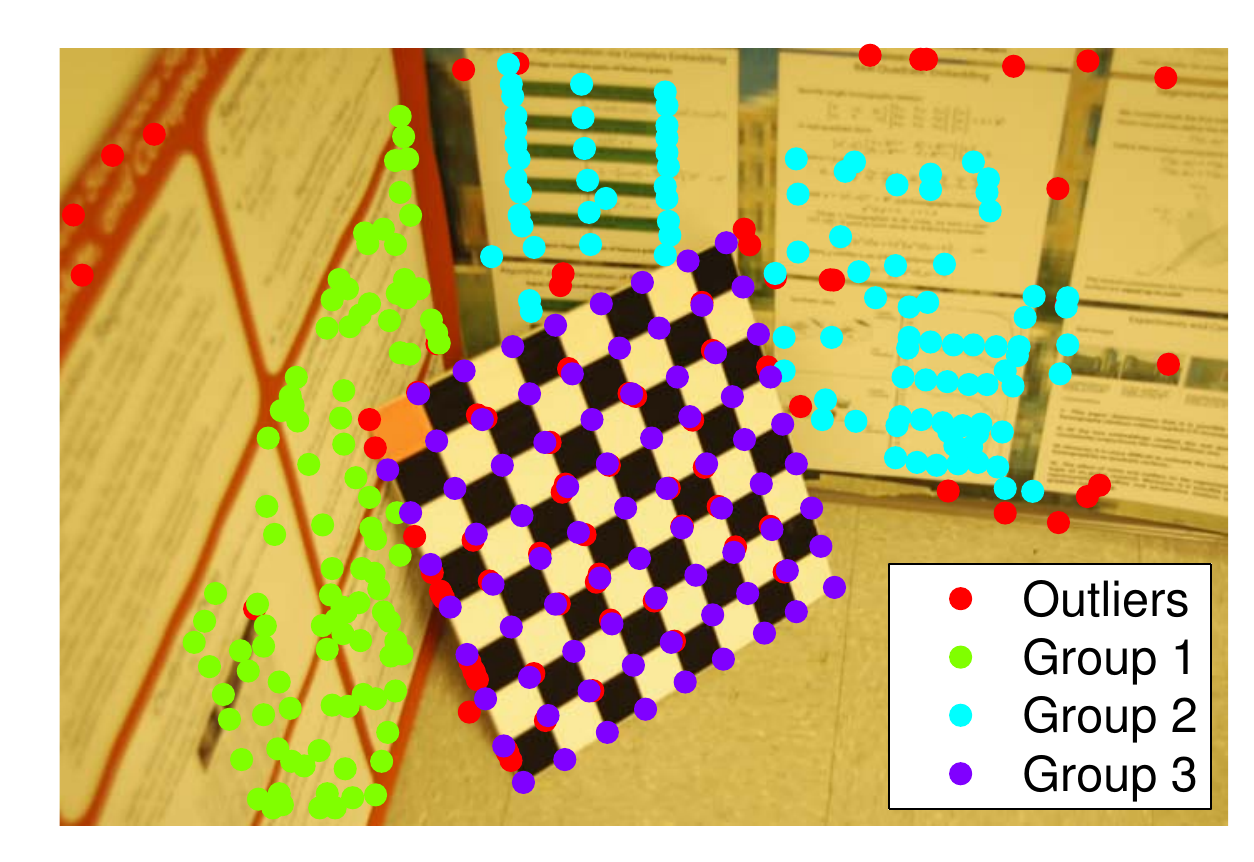}
		\label{fig:posterCheckerboard:subfig4}
	}
	
	\subfloat[]{
		\includegraphics[height=2.0in,width=2.2in] {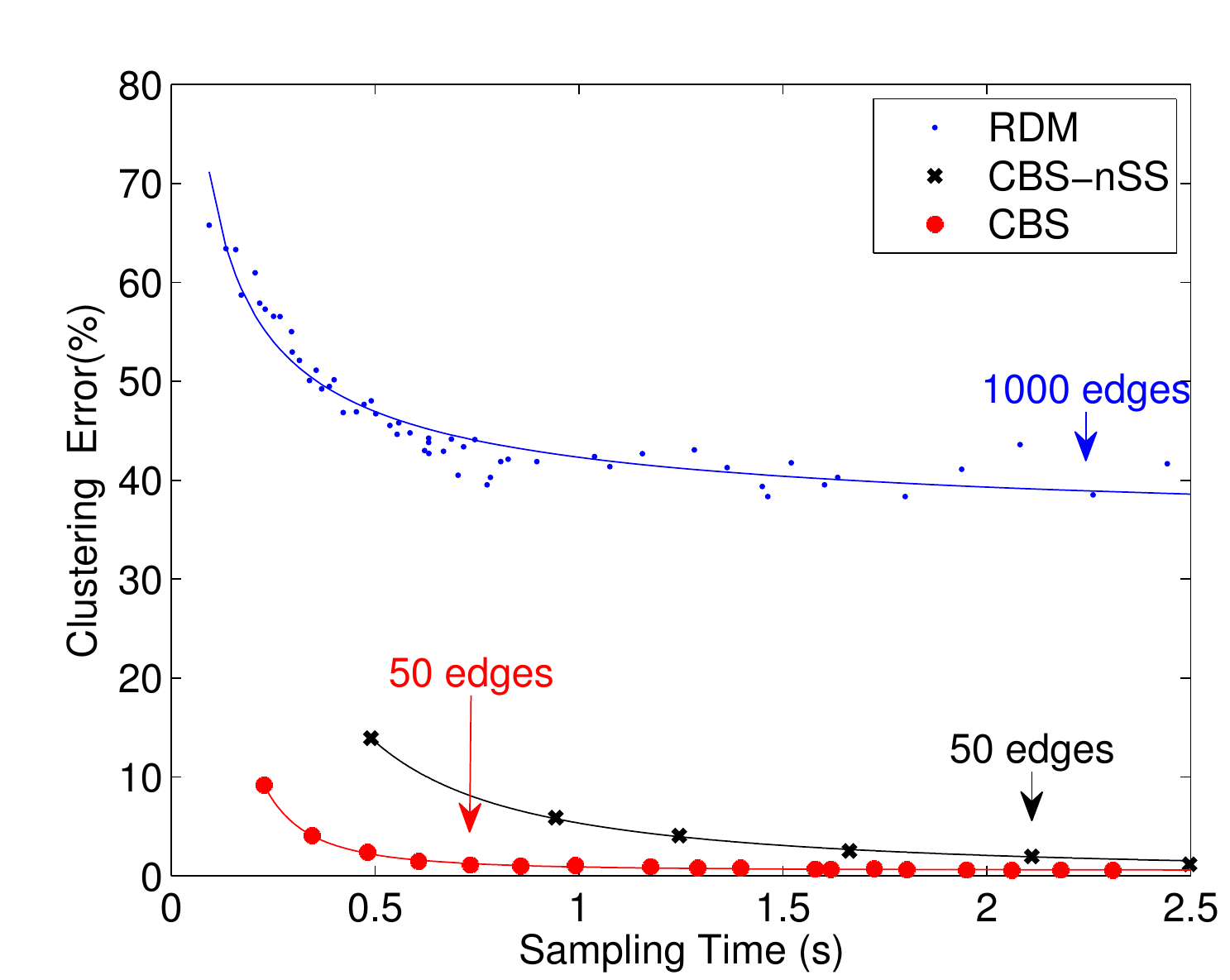}
		\label{fig:posterCheckerboard:subfig5}
	}
	\subfloat[]{
		\includegraphics[height=1.9in,width=2.2in] {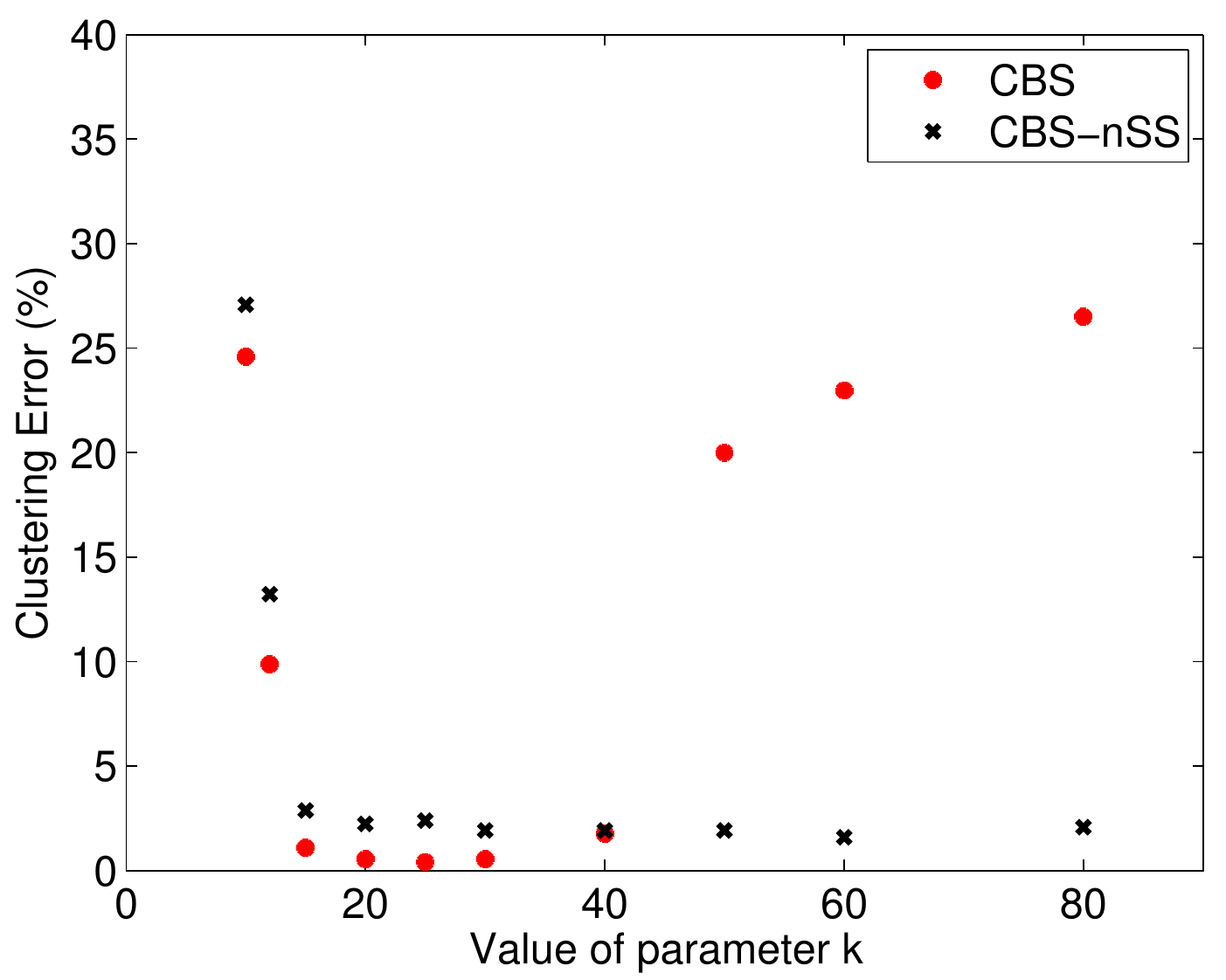}
		\label{fig:posterCheckerboard:subfig6}
	}
	\subfloat[]{
		\includegraphics[height=2.0in,width=2.2in] {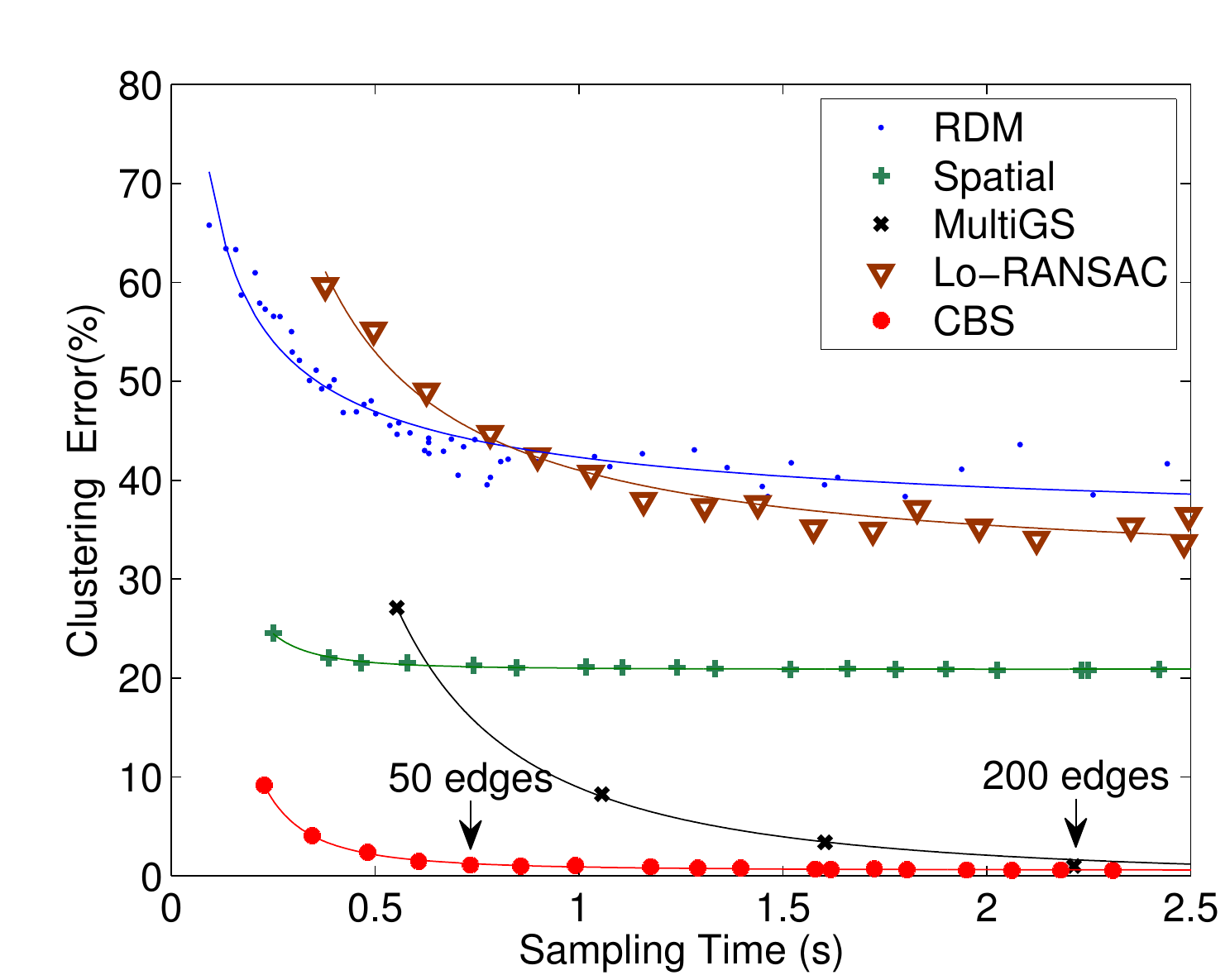}
		\label{fig:posterCheckerboard:subfig7}
	}

	\caption{The results on ``\textit{posters-checkerboard}'' sequence, \ref{fig:posterCheckerboard:subfig1} shows the ground truth clustering while  \ref{fig:posterCheckerboard:subfig2} - \ref{fig:posterCheckerboard:subfig4} shows the clustering obtained with RDM, CBS-nss and CBS at 1s. \ref{fig:posterCheckerboard:subfig5} and \ref{fig:posterCheckerboard:subfig6} shows the variation of clustering error with time and the value of parameter $k$ respectively, while \ref{fig:posterCheckerboard:subfig7} shows the variation in clustering error with the value of parameter $k$ (best viewed in color).}
	\label{fig:posterCheckerboard}
\end{figure*}

We used the ``\textit{posters-checkerboard}'' sequence from RAS dataset \cite{Rao2010} to evaluate the significance of the main components of the CBS method. This sequence contain three rigid moving objects with 100, 99, 81 point matches respectively and 99 outlier points. 
In the first experiment the matrix $H$ was generated with edges obtained by: pure random sampling (RDM), with the CBS method without the sub-sampling strategy, i.e. lines 3, 7-10 removed from Algotihm~\ref{alg:PM} (CBS-nSS) and the complete proposed method (CBS) respectively. For each sampling method the number of hypothesis ($n_H$) was varied and the mean clustering error and the run time was recorded (averaged over 100 runs per each $n_H$). \figref{posterCheckerboard:subfig5} shows the variation of mean clustering error with the sampling time (computing time). The results show that for this problem accurate identification of models could not be achieved with pure random sampling even when large number of edges were sampled. It also shows that the sub-sampling strategy of the proposed CBS method significantly contributes towards accurate and efficient identification of the underlying models in data. 

Next we use the same image sequence to study the variations in accuracy of the proposed method with the value of parameter $k$.  This parameter defines the minimal acceptable size for a structure (in number of points) in a given application. Here we vary the value of $k$ from 10 to 80 (CBS use edge of size 10 and the smallest structure in this sequence has only 81 points hence any value outside this range is not realistic). The number of hypothesis was set to 100 for both sampling methods.
Results plotted in \figref{posterCheckerboard:subfig6} show that for CBS-nSS and CBS the clustering error reduces steeply up to around $k=20$. In CBS-nSS the CE remains relatively unchanged after that while in CBS the clustering error start to increase when $k$ goes beyond $40$. This behavior can be explained as follows: The CBS method estimates the scale of noise from  data and the analysis of \cite{Hoseinnezhad2010} showed that the estimation of the noise scale from data requires \textit{at least} $20$ data points to limit the effects of finite sample bias. As such, the CBS method would not have high accuracy when $k<20$. In addition the data sub-sampling in CBS reduces the number of points available for each run of the sample generator hence the increased clustering error for large $k$ values. Using large values for $k$ is also not desirable because the smaller size structures would be ignored. 

Next, we compared the proposed hypothesis generation process against several well known sampling methods for robust model fitting (e.g. MultiGS \cite{Tat-Jun2012} and Lo-RANSAC~\cite{Chum2003}). These methods are designed to bias the sampling process towards selecting points from a structure in data. 
For completeness we have also included pure spatial sampling (generate hypothesis using points closer in space picked via a KDtree) and random sampling. Similar to the proposed method the hypothesis from these sampling methods were used to generate a graph which is cut to perform the clustering. The \figref{posterCheckerboard:subfig6} shows that the CBS method is capable of generating highly accurate clusterings faster than other sampling methods.

It should be noted here that while we have only presented the results for one two-view motion segmentation case, similar trends were observed across all other problems tested in this paper.

\subsection{Two-view motion segmentation}
\label{sec:twoviewMS}
Two-view motion segmentation is the task of identifying the points correspondences of each object in two views of a dynamic scene that contains multiple independently moving objects. Provided that the point matches between the two views are given as
$[X_1, X_2]$ where $X_i = (x, y, 1)^\top$ is a coordinate of a point in view $i$, each motion can be modeled using the fundamental matrix $F \in \mathcal{R}^{3 \times 3}$  as \cite{Torr1997}:
\begin{equation}
	X_1^\top F X_2 = 0
\end{equation}
The distance from a given model to a point pair can be measured using the Sampson distance \cite{Hartley2003}.

We tested the performance of the CBS method on the Adelaide-RMF dataset \cite{HoiSim2011} which contains key-point matches (obtained using SIFT) of dynamic scenes together with the ground truth clustering. The clustering error and the computational time of the CBS method on each sequence together with those of the competing methods (PEARL, FLOSS, RCMSA and QP-MF) are given in \tabref{fundamentalRes}. The results show that in comparison to the competing methods, the proposed method has achieved comparable or better accuracy over all sequences. Moreover, on average the computation time of the proposed method is around 4 times less than that of QP-MF and twice that of the RCMSA when its computational bottlenecks are implemented using C (MATLAB MEX) whereas our method is implemented using simple MATLAB script. One would expect significant improvements in terms of speed by using C language implementation.

In these experiments the parameter $k$ of the proposed method was set to $k = min(0.1 \times N , 20)$. The number of samples in QP-MF was set to 200 (determined through trial and error: no significant improvement of accuracy was observed when the  number of samples were increased beyond 200 for a test sequence).

\begin{table*}[htbp]
	\centering
	\caption{Two-view motion segmentation results on Adelaide-RMF dataset. The Median CE values of PEARL and FLOSS \cite{Lazic2009} repoted in \cite{Pham2014} are used here.}
	\begin{tabular}{|r|c|c|c|c|c|c|c|c|}
		\hline
		\multicolumn{1}{|c|}{\multirow{3}[4]{*}{}} & \multirow{2}[2]{*}{PEARL} & \multirow{2}[2]{*}{FLOSS} & \multicolumn{2}{c|}{\multirow{2}[2]{*}{\textit{QP-MF}}} & \multicolumn{2}{c|}{\multirow{2}[2]{*}{\textit{RCMSA}}} & \multicolumn{2}{c|}{\multirow{2}[2]{*}{\textit{CBS}}} \\
		\multicolumn{1}{|c|}{} &       &       & \multicolumn{2}{c|}{} & \multicolumn{2}{c|}{} & \multicolumn{2}{c|}{} \\
		\cline{2-9}    \multicolumn{1}{|c|}{} & \textit{Median CE} & \textit{Median CE} & \textit{Median CE} & \textit{Time} & \textit{Median CE} & \textit{Time} & \textit{Median CE} & \textit{Time} \\
		\hline
		\textit{biscuitbookbox} & {8.11} & 11.58 & 5.02  & 4.78  & 7.72  & 0.56  & \textbf{0.00} & 0.95 \\
		\hline
		\textit{boardgame} & {16.85} & 17.92 & 17.38 & 4.49  & 12.09 & 0.50  & \textbf{11.28} & 0.99 \\
		\hline
		\textit{breadcartoychips} & {12.24} & 15.82 & 8.65  & 4.52  & 9.97  & 0.64  & \textbf{5.63} & 0.93 \\
		\hline
		\textit{breadcubechips} & {9.57} & 11.74 & 3.04  & 4.47  & 9.78  & 0.54  & \textbf{0.87} & 0.85 \\
		\hline
		\textit{breadtoycar} & {10.24} & 11.75 & 6.33  & 4.20  & 8.73  & 0.44  & \textbf{3.96} & 0.75 \\
		\hline
		\textit{carchipscube} & {10.30} & 16.97 & 17.27 & 3.59  & 4.85  & 0.42  & \textbf{2.44} & 0.65 \\
		\hline
		\textit{cubebreadtoychips} & {9.02} & 11.31 & 2.14  & 5.07  & 8.87  & 0.71  & \textbf{1.91} & 1.13 \\
		\hline
		\textit{dinobooks} & {19.17} & 20.28 & 17.92 & 5.20  & 17.50 & 0.73  & \textbf{12.98} & 1.25 \\
		\hline
		\textit{toycubecar} & {12.00} & 13.75 & 14.50 & 3.71  & \textbf{11.00} & 0.38  & 19.19 & 0.70 \\
		\hline
	\end{tabular}%
	\label{tab:fundamentalRes}%
\end{table*}%

\subsection{3D-motion segmentation of rigid bodies}
The objective of 3D motion segmentation is to identify multiple moving objects using point trajectories through a video sequence. If the projections (to the image plane) of $N$ points tracked through $F$  frames are available, $ \left [  x_{f\alpha} \right ]_{\alpha =1 \dots N}^{f=1 \dots F}: x_{f\alpha} \in \mathcal{R}^2$ then \cite{Sugaya2004} has shown that the point trajectories $ P_\alpha = \left [ x_{1\alpha}, y_{1\alpha},x_{2\alpha}, \dots x_{F\alpha}, y_{F\alpha}  \right ]^\top \in \mathcal{R}^{2F} $ that belong to a single rigid moving object are contained within a subspace of $rank \leq 4$, under the affine camera projection model. Hence, the problem of 3D motion segmentation can be reduced to a subspace clustering problem. 

One of the characteristics in subspace segmentation is that the dimension of the subspaces may vary between two and four, depending on the nature of the motions. This means that the model we are estimating is not fixed. The proposed method, which was not specifically developed to solve this problem (unlike some competing techniques \cite{Elhamifar2013}) is not capable of identifying the number of dimensions of a given motion and requires this information as an input. In our implementation we have used the Eigan values of the sampled data point to select a dimension $d$ of the model such that $2 \leq d \leq 4$.

We utilized the commonly used ``checkerboard'' image sequence in the Hopkins 155 dataset \cite{Tron2007} to evaluate the CBS algorithm. This dataset contains trajectory information of 104 video sequences that are categorized into two main groups depending on the number of motions in each sequence (two or three motions). 

\begin{figure*}[!t]
	\centering
	\subfloat[Two motion sequences]{
		\includegraphics[height=2.0in,width=2.25in] {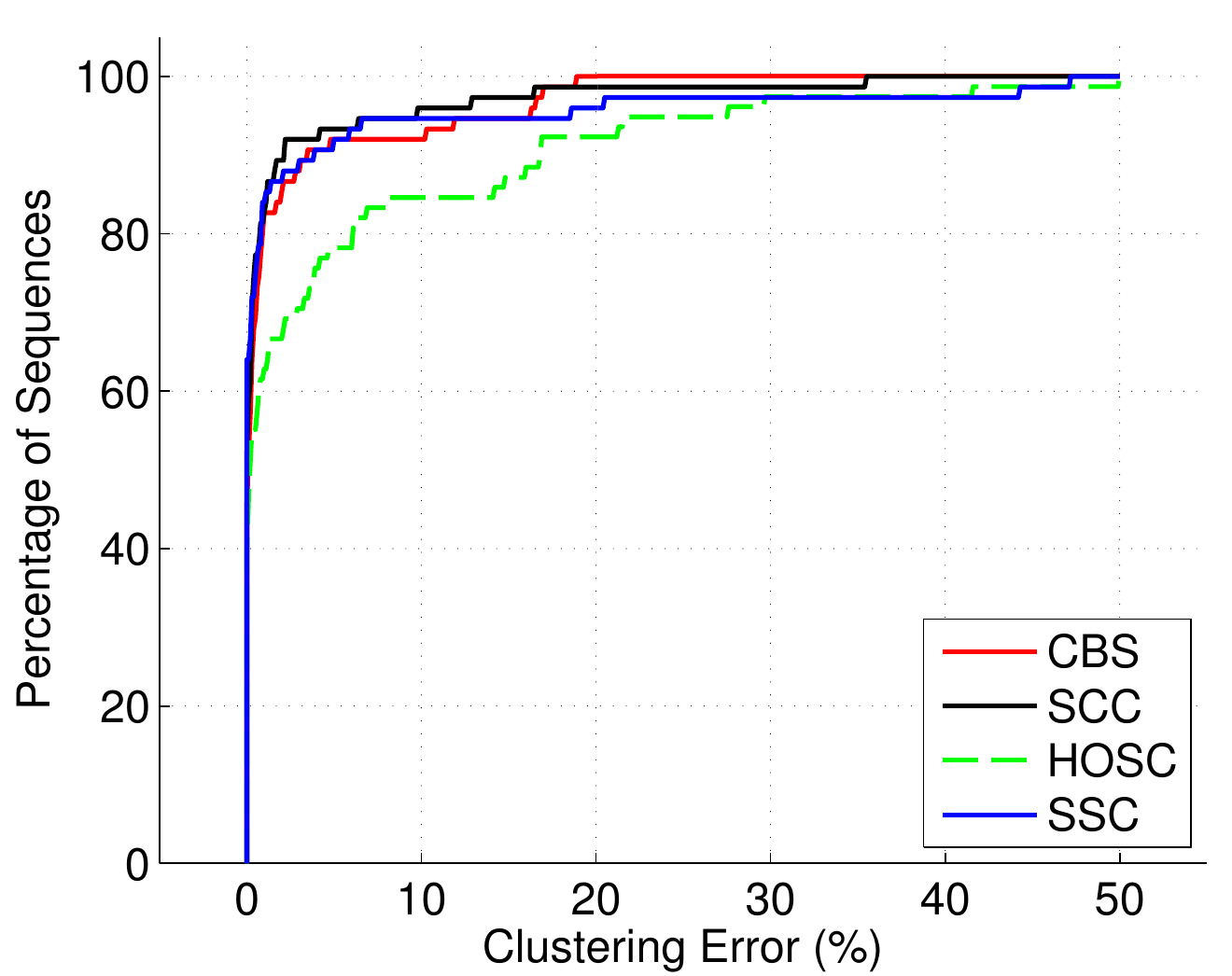}
		\label{fig:hopkinHist:subfig1}
	}
	\subfloat[Three motion sequences]{
		\includegraphics[height=2.0in,width=2.25in] {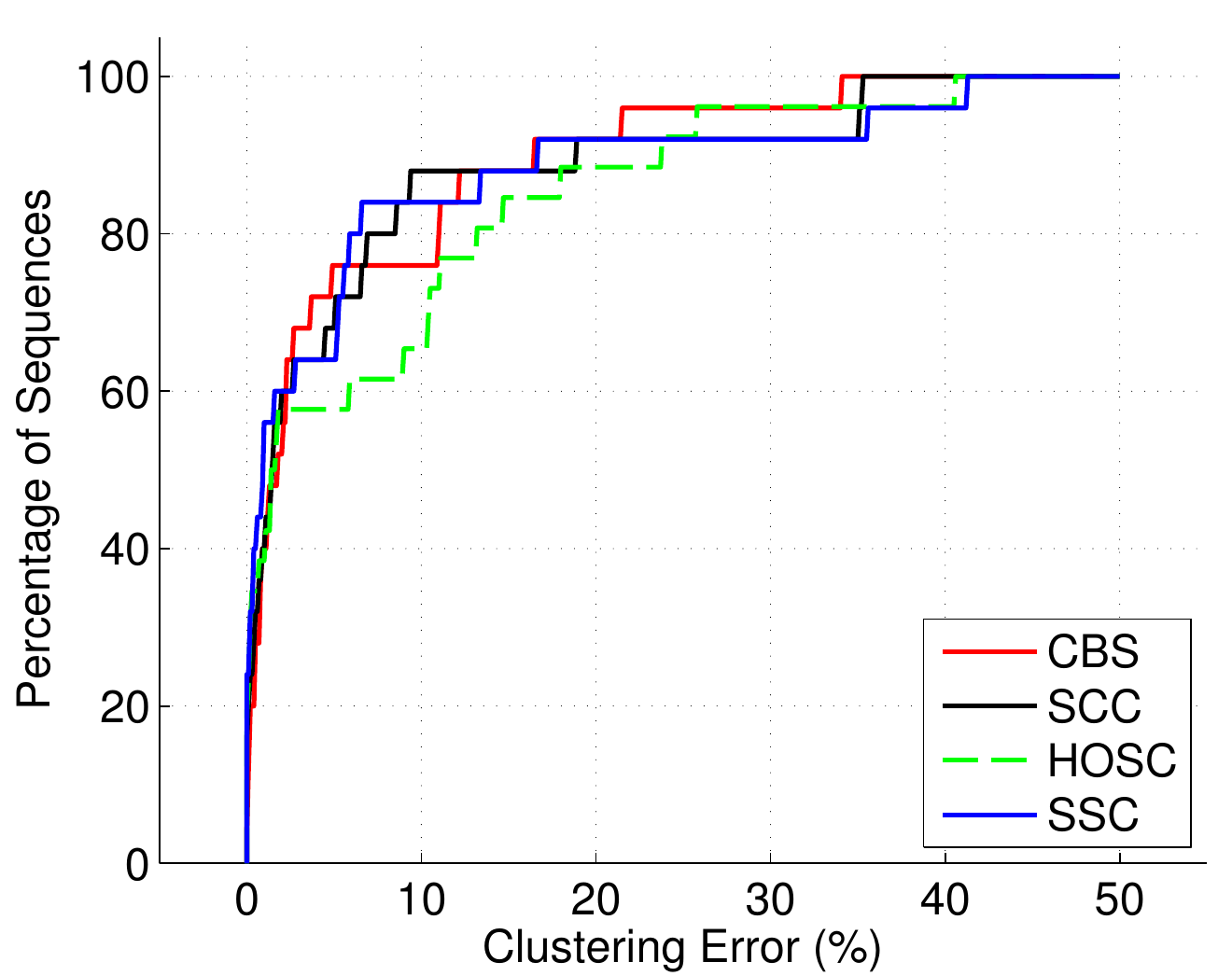}
		\label{fig:hopkinHist:subfig2}
	}
	\subfloat[Three motion with Outliers]{
		\includegraphics[height=2.0in,width=2.25in] {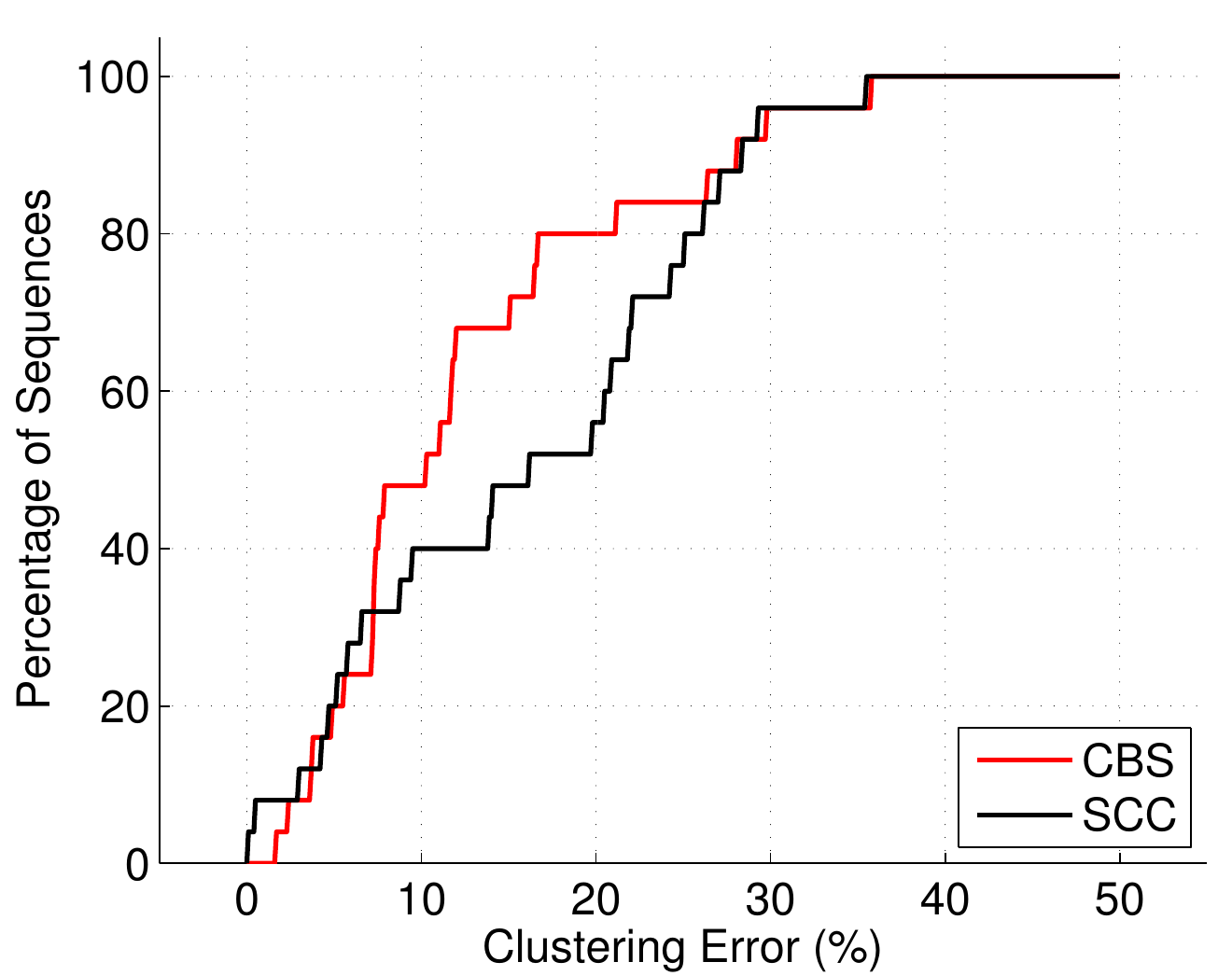}
		\label{fig:hopkinHist:subfig3}
	}
	
	\caption{Cumulative distributions of the clustering errors (CE) per sequence of the Hopkings dataset. \figref{hopkinHist:subfig1} Two motion sequences, \figref{hopkinHist:subfig2} Three motion sequences and \figref{hopkinHist:subfig3} Three motion sequences with added synthetic outliers. }
	\label{fig:hopkinHist}
\end{figure*}

The clustering error (mean and median) and the computation time for CBS together with competing higher order affinity based methods are shown in \tabref{hopkins155}. The results show that CBS has achieved comparable clustering accuracies to those achieved by competing methods while being significantly faster than those methods (specially on 3-motion sequence). For completeness we have also included the results for some energy minimization (PEARL\cite{Boykov2001}, QP-MF\cite{Jin2011}) and fit \& remove (RANSAC, HMSS\cite{Tennakoon2015}) based methods as reported in \cite{Jin2011}. To gain a better understanding of the methods (that has good accuracy) across all sequences we have plotted the cumulative distributions of the errors per sequence in \figref{hopkinHist:subfig1} (two motion sequences)  and \figref{hopkinHist:subfig2} (three motion sequences). 
For algorithms with a random elements the mean error across 100 runs is used.

\begin{figure*}[!t]
	\centering
	\subfloat[three-cars]{
		\includegraphics[height=1.5in,width=1.6in] {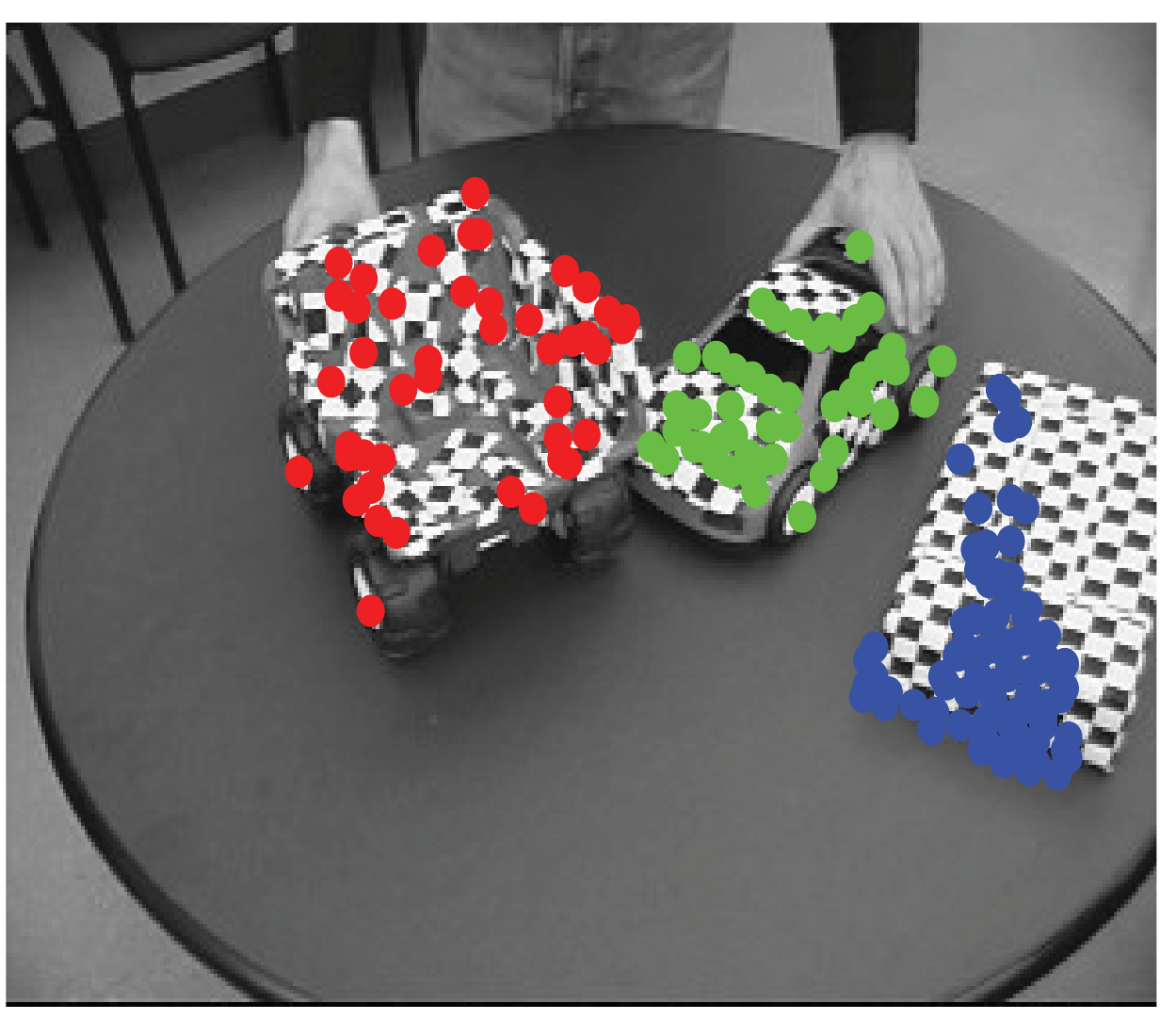}
		\label{fig:hopkinQuality:subfig1}
	}
	\subfloat[articulated]{
		\includegraphics[height=1.5in,width=1.6in] {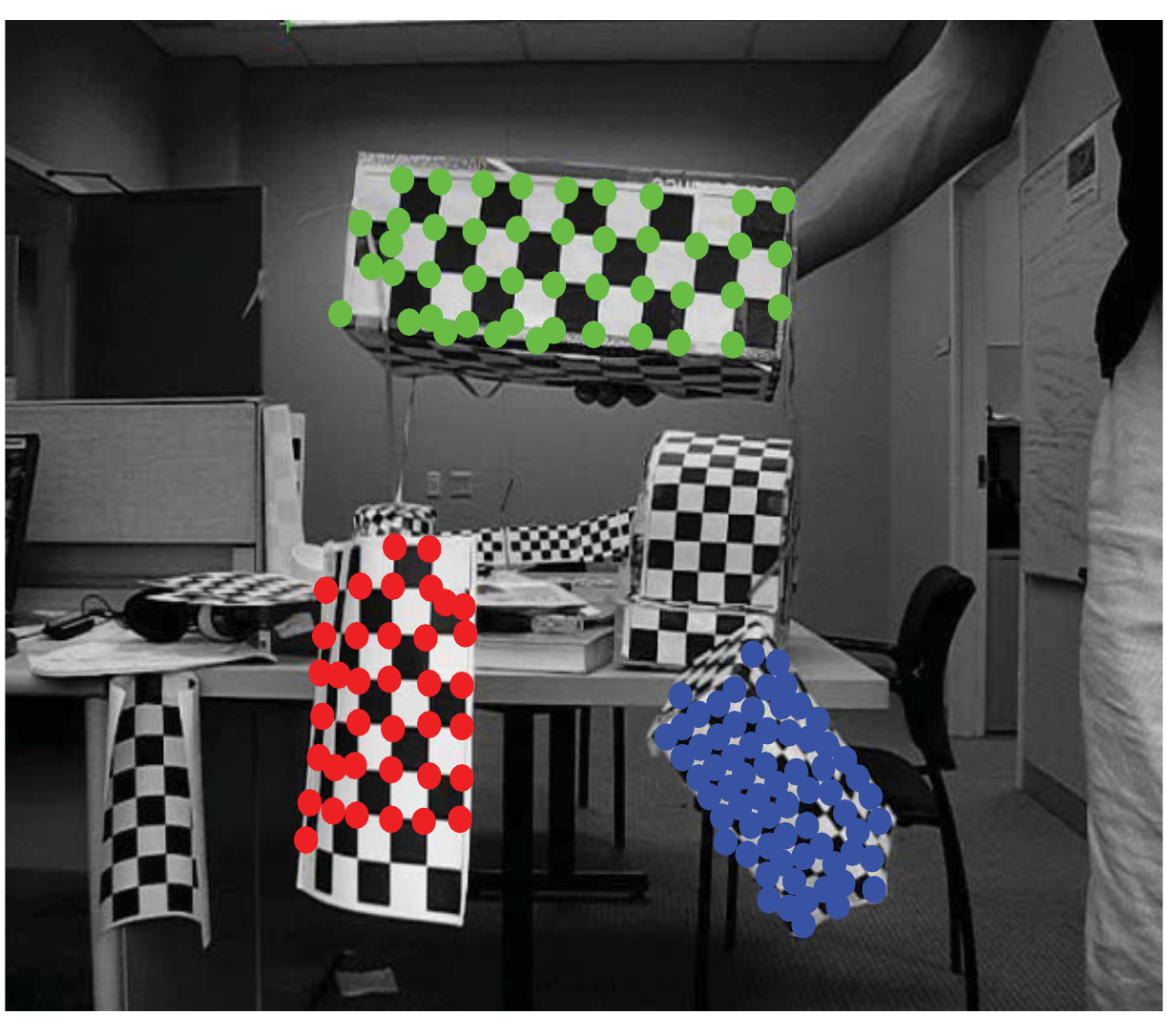}
		\label{fig:hopkinQuality:subfig2}
	}
	\subfloat[cars5]{
		\includegraphics[height=1.5in,width=1.6in] {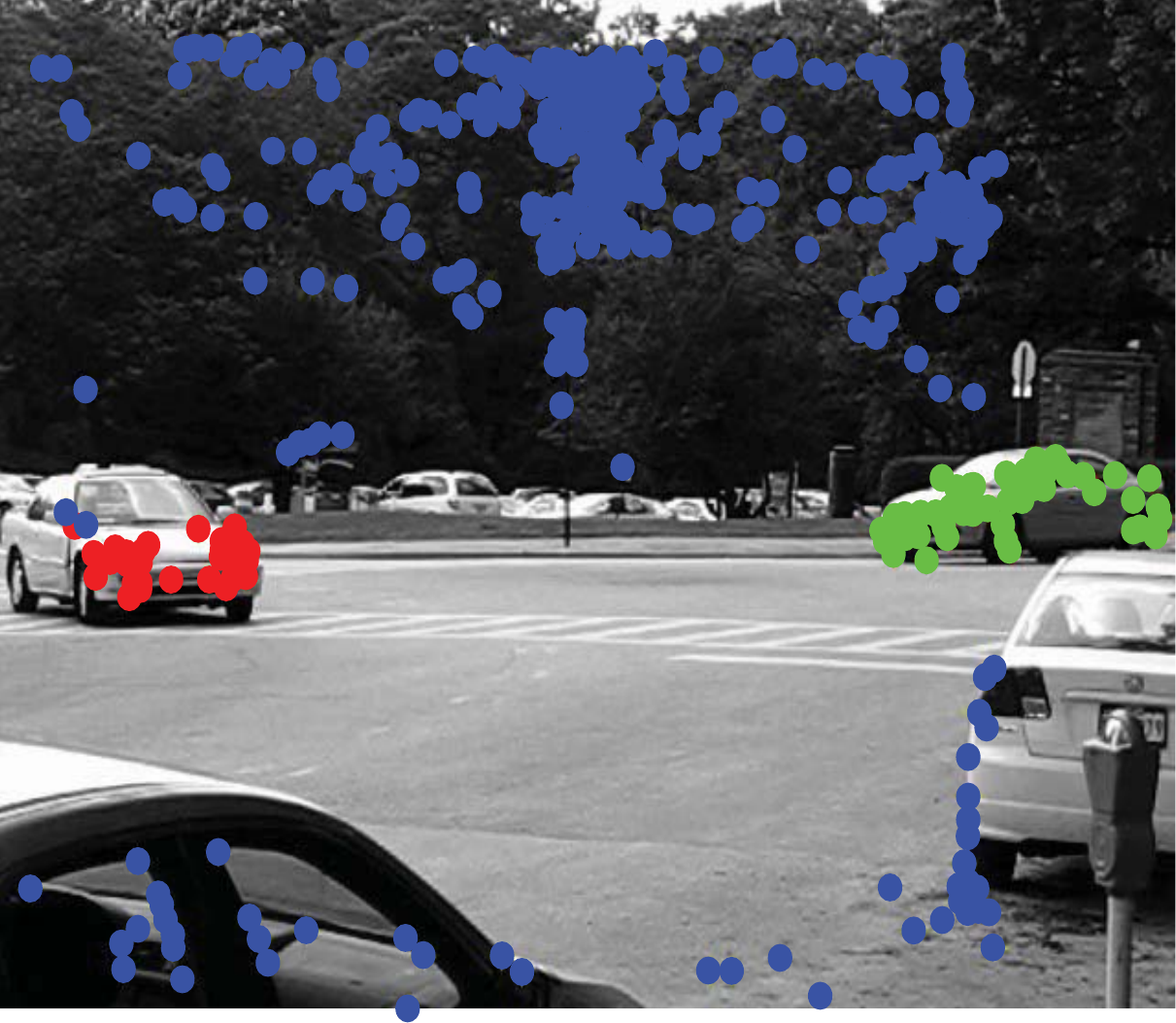}
		\label{fig:hopkinQuality:subfig3}
	}
	\subfloat[cars2-07]{
		\includegraphics[height=1.5in,width=1.6in] {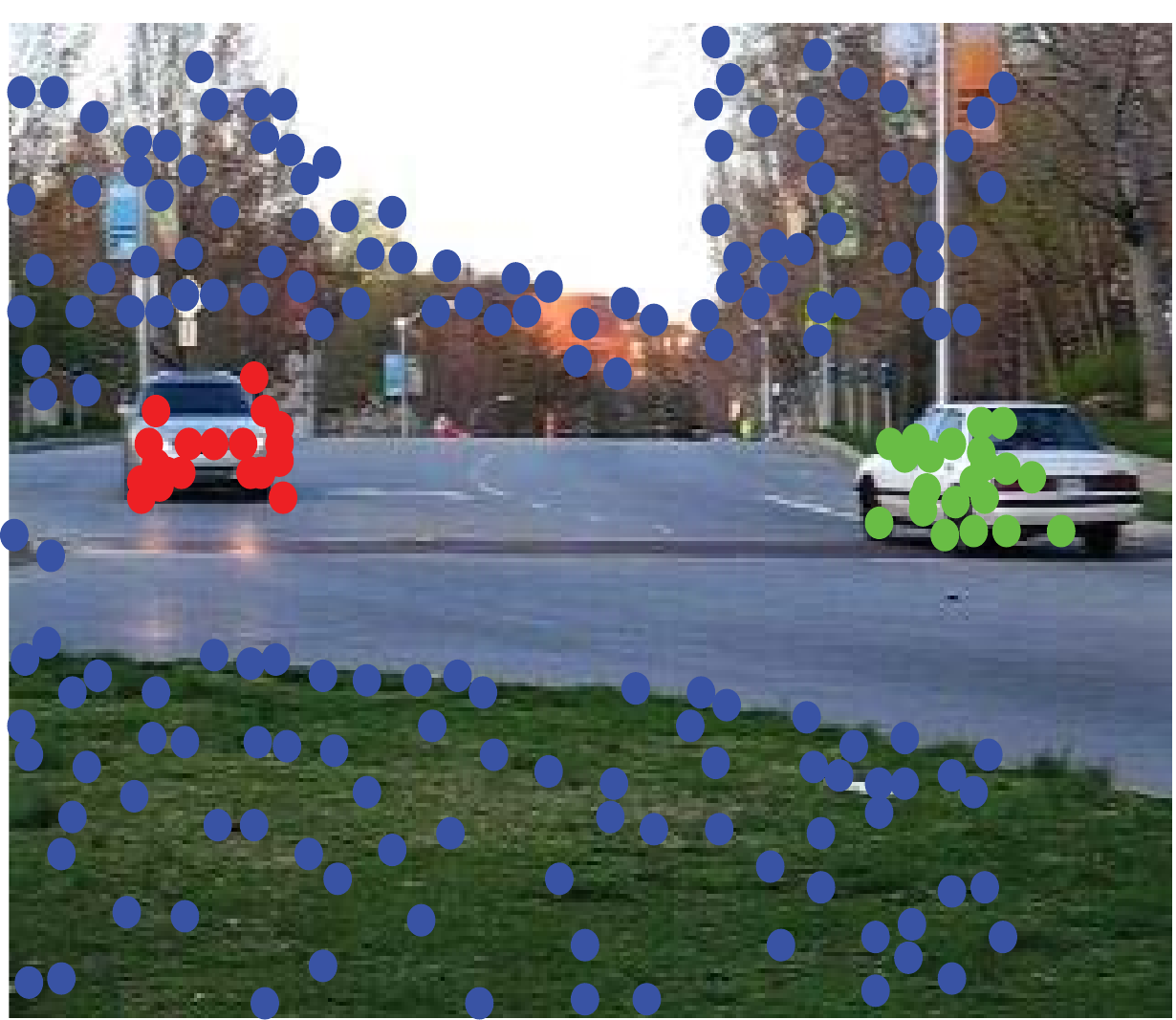}
		\label{fig:hopkinQuality:subfig4}
	}

	\subfloat[1RT2RTCRT-B]{
		\includegraphics[height=1.5in,width=1.6in] {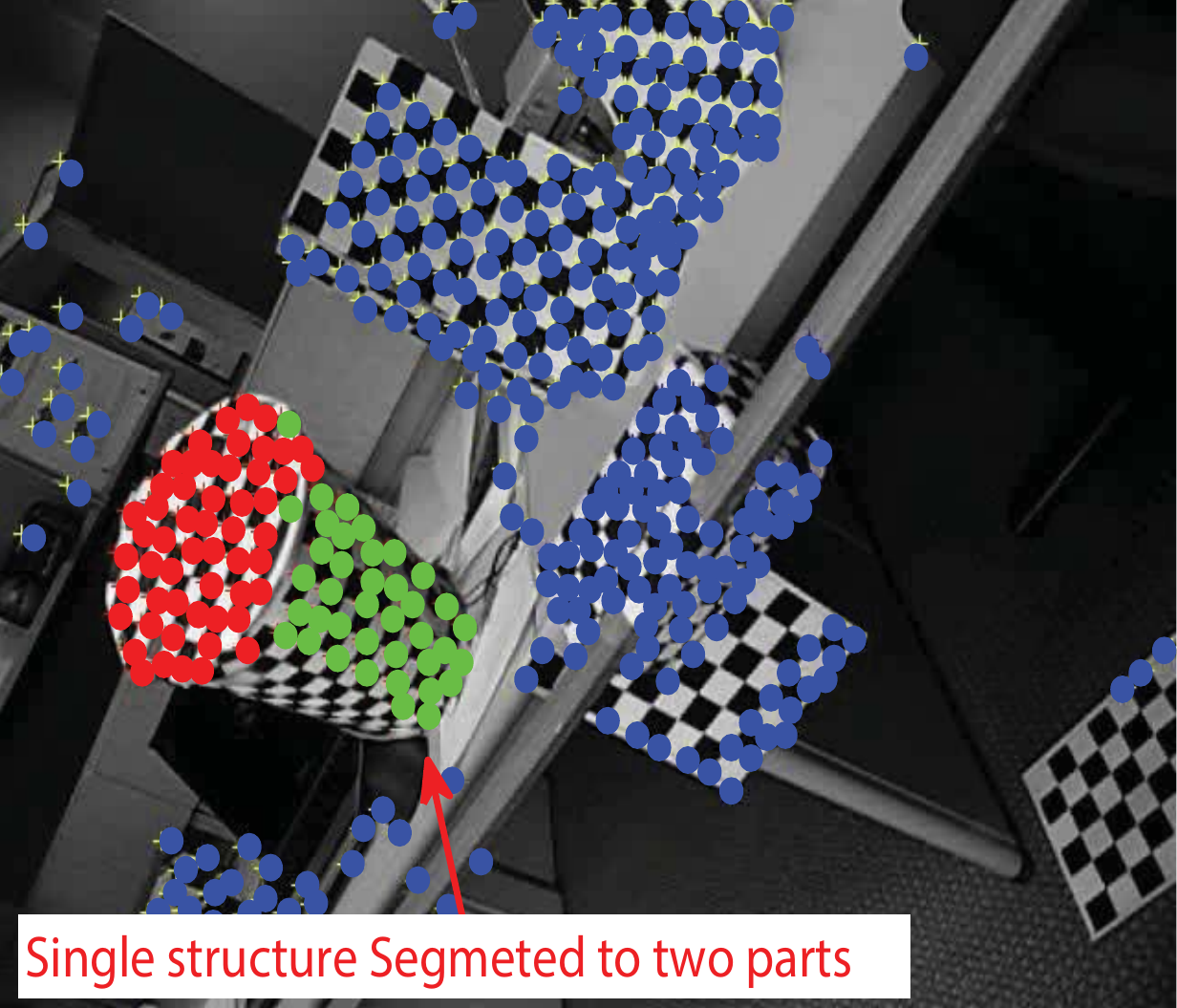}
		\label{fig:hopkinQuality:subfig5}
	}
	\subfloat[2RT3RCR]{
		\includegraphics[height=1.5in,width=1.6in] {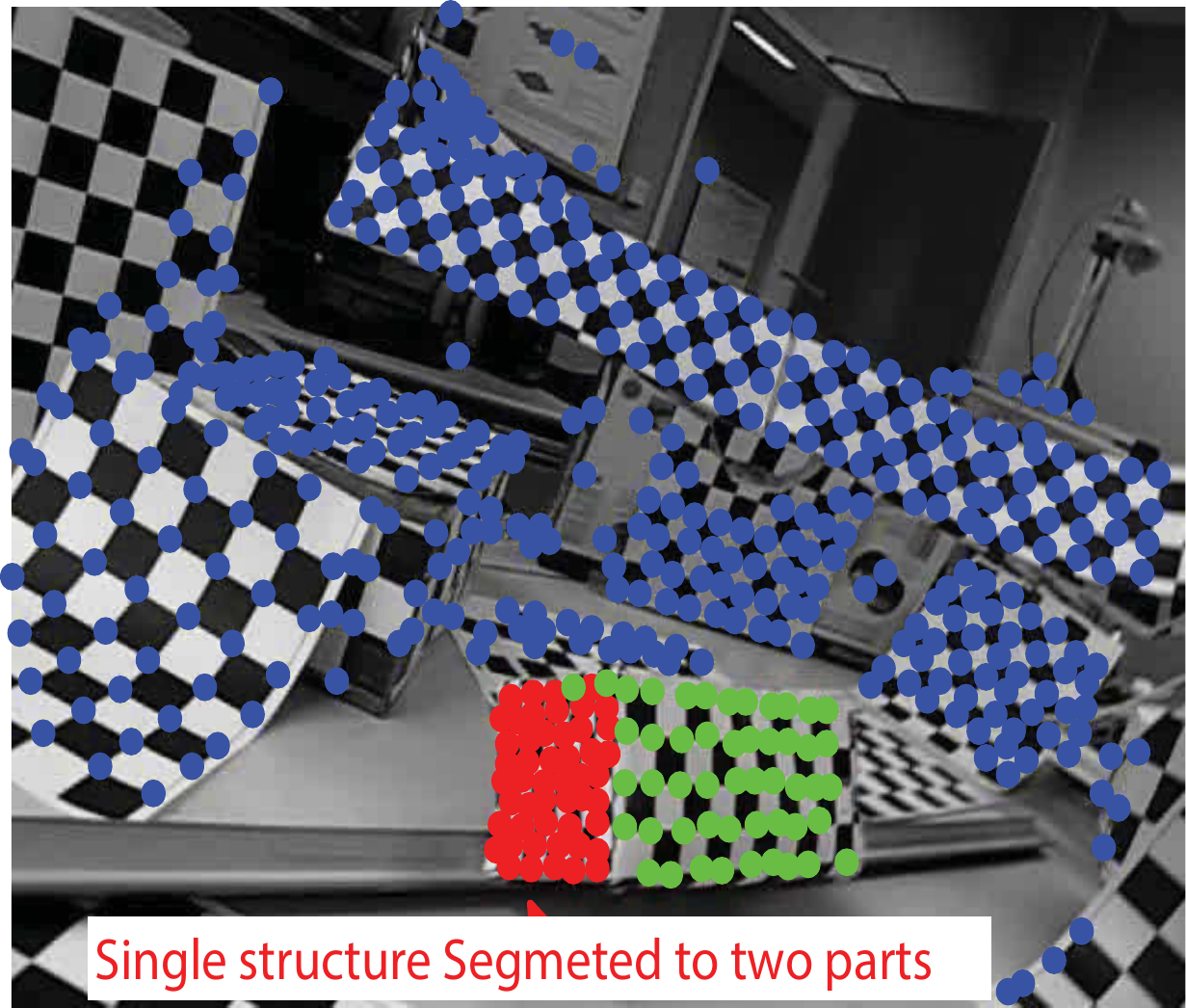}
		\label{fig:hopkinQuality:subfig6}
	}
	\subfloat[cars9]{
		\includegraphics[height=1.5in,width=1.6in] {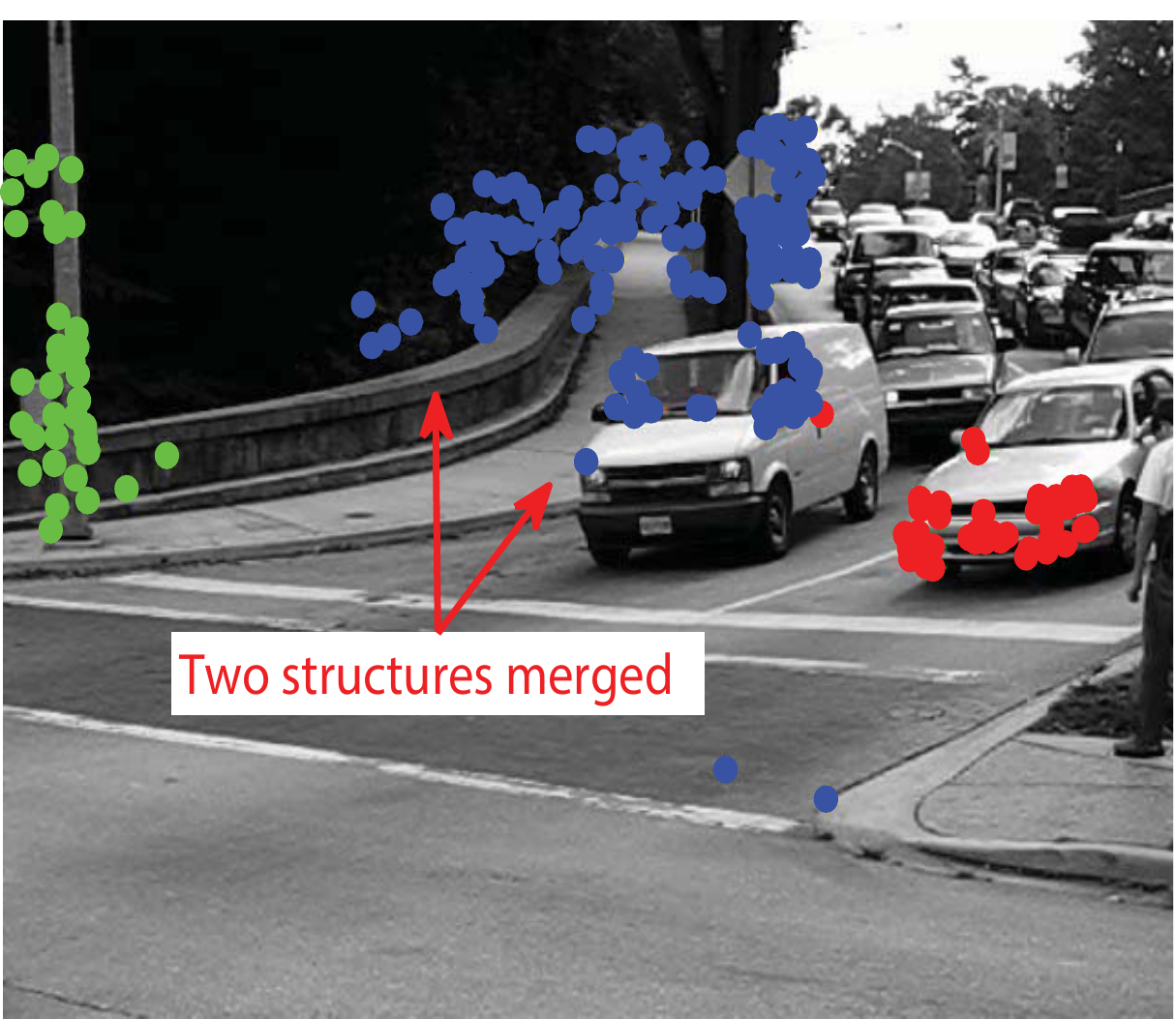}
		\label{fig:hopkinQuality:subfig7}
	}
	\subfloat[cars2B]{
		\includegraphics[height=1.5in,width=1.6in] {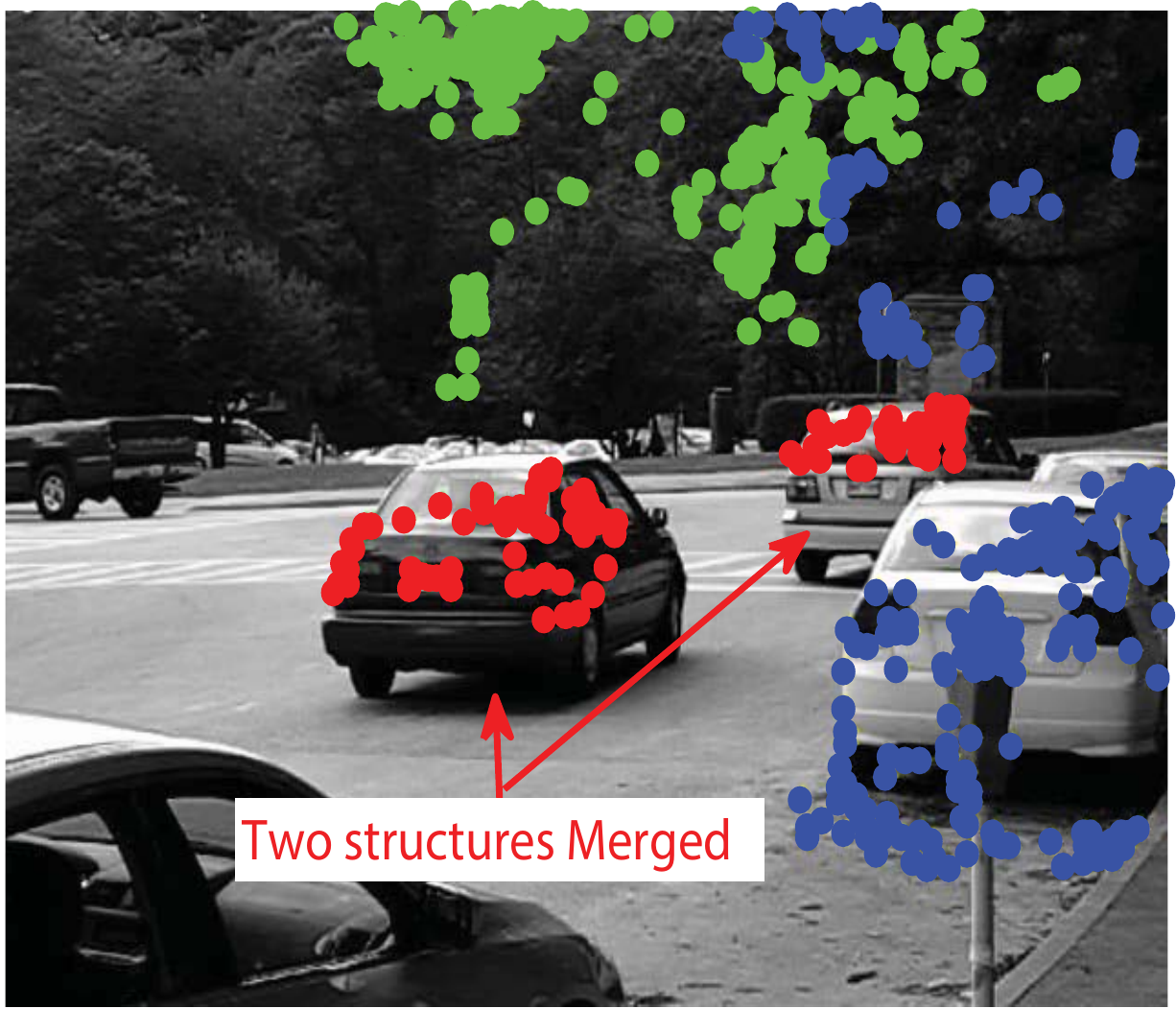}
		\label{fig:hopkinQuality:subfig8}
	}
	
	\caption{Clustering results obtained using the proposed method on several examples sequences from the Hopkings dataset. The top row show cases where the proposed method has been sucessful whereas the bottom row show cases where the proposed method failed to identify all the clusters correctly (best viewed in colour). }
	\label{fig:hopkinQuality}
\end{figure*}

To provide a qualitative measure of the performance the final segmentation results of several sequences in the Hopkins 155 dataset, where CBS was both successful and unsucessful, are shown in \figref{hopkinQuality}.

The sequences contained in the Hopkins 155 dataset are outlier-free. In order to test robustness to outliers, we added synthetically generated outlier trajectories to each three-motion sequence of Hopkins 155 dataset\footnote{The MATLAB code provided by http://www.vision.jhu.edu/data/hopkins155/ was used.}. The clustering results of the CBS method together with those obtained by the best performing method (SCC) are plotted in \figref{hopkinHist:subfig3}. The results show that CBS was able to achieve high accuracy in the presence of outliers on higher number of sequences. It should be noted here that the SSC algorithm is not designed to handle outliers and therefore was not included in this analysis.

\begin{table*}[htbp]
	\centering
	\caption{Comparative performance in terms of accuracy and speed using Hopkings 155 checkerboard sequence.}
	\begin{tabular}{|r|r|r|r|r|r|r|r|r|}
		\hline
		\multicolumn{1}{|c|}{\multirow{2}[4]{*}{}} & \multicolumn{1}{c|}{RANSAC} & \multicolumn{1}{c|}{PEARL} & \multicolumn{1}{c|}{QP-MF} & \multicolumn{1}{c|}{HMSS} & \multicolumn{1}{c|}{SSC\textsuperscript{*}} & \multicolumn{1}{c|}{SCC} & \multicolumn{1}{c|}{HOSC} & \multicolumn{1}{c|}{CBS} \\
		\cline{2-9}    \multicolumn{1}{|c|}{} & \multicolumn{8}{c|}{\textit{2 Motion Sequences}} \\
		\hline
		Mean  & \multicolumn{1}{c|}{6.52} & \multicolumn{1}{c|}{5.28} & \multicolumn{1}{c|}{9.98} & \multicolumn{1}{c|}{3.98} & \multicolumn{1}{c|}{2.23} & \multicolumn{1}{c|}{\textbf{1.40}} & \multicolumn{1}{c|}{5.28} & \multicolumn{1}{c|}{1.60} \\
		\hline
		Median & \multicolumn{1}{c|}{1.75} & \multicolumn{1}{c|}{1.83} & \multicolumn{1}{c|}{1.38} & \multicolumn{1}{c|}{0.00} & \multicolumn{1}{c|}{0.00} & \multicolumn{1}{c|}{\textbf{0.04}} & \multicolumn{1}{c|}{0.02} & \multicolumn{1}{c|}{0.10} \\
		\hline
		Time  & \multicolumn{1}{c|}{-} & \multicolumn{1}{c|}{-} & \multicolumn{1}{c|}{-} & \multicolumn{1}{c|}{-} & \multicolumn{1}{c|}{0.65} & \multicolumn{1}{c|}{0.66} & \multicolumn{1}{c|}{1.27} & \multicolumn{1}{c|}{\textbf{0.48}} \\
		\hline
		& \multicolumn{8}{c|}{\textit{3 Motion Sequences}} \\
		\hline
		Mean  & \multicolumn{1}{c|}{25.78} & \multicolumn{1}{c|}{21.38} & \multicolumn{1}{c|}{15.61} & \multicolumn{1}{c|}{11.06} & \multicolumn{1}{c|}{5.77} & \multicolumn{1}{c|}{5.74} & \multicolumn{1}{c|}{7.38} & \multicolumn{1}{c|}{\textbf{4.98}} \\
		\hline
		Median & \multicolumn{1}{c|}{26.01} & \multicolumn{1}{c|}{21.14} & \multicolumn{1}{c|}{8.82} & \multicolumn{1}{c|}{1.20} & \multicolumn{1}{c|}{\textbf{0.95}} & \multicolumn{1}{c|}{1.48} & \multicolumn{1}{c|}{1.53} & \multicolumn{1}{c|}{1.04} \\
		\hline
		Time  & \multicolumn{1}{c|}{-} & \multicolumn{1}{c|}{-} & \multicolumn{1}{c|}{-} & \multicolumn{1}{c|}{-} & \multicolumn{1}{c|}{1.47} & \multicolumn{1}{c|}{1.29} & \multicolumn{1}{c|}{2.00} & \multicolumn{1}{c|}{\textbf{0.55}} \\
		\hline
		\multicolumn{9}{l}{\textsuperscript{*}\footnotesize{The results for SSC are generated using the faster ADMM \cite{Elhamifar2013} implementation provided}}\\ \multicolumn{9}{l}{\footnotesize{in http://vision.jhu.edu/ without any modifications. The SSC CSX implementation \cite{Elhamifar2009} }}\\
		\multicolumn{9}{l}{\footnotesize{is more accurate but has significantly higher computational cost.}} \\
	\end{tabular}%
	\label{tab:hopkins155}%
\end{table*}%

\subsection{Long-term analysis of moving objects in video}

The point trajectories of the ``Hopkings155'' dataset used in the above analysis are hand tunned (i.e. the point trajectories of each sequence are cleaned by a human such that they do not contain gross-outliers or incomplete trajectories).  Recently, more realistic ``Berkeley Motion Segmentation Dataset'' (BMS-26) was introduced by \cite{Ochs2014}, \cite{Brox2010} for long-term analysis of moving objects in video. This dataset consist of point trajectories that are obtained by running a state of the art feature point tracker (the large displacement optical flow \cite{Brox2011}), on 26 videos directly without any further post processing.  Thus those feature trajectories contain noise and outliers and most importantly include incomplete trajectories. Incomplete trajectories are trajectories that do not run for the whole duration of the video, they can appear in any frame of the video and disappear on or before the last frame. These incomplete trajectories are mainly caused by occlusion and disocclusion.

The traditional approach of using two views to segment objects is susceptible to short term variations (e.g. human standing for a short time can be merged with the background). Hence Brox and Malik \cite{Brox2010} proposed long-term video analysis where a similarity between two points trajectories was used to build a graph that was segmented using spectral clustering. Such pairwise affinities only model translations and do not account for scaling and rotation. Ochs and Brox \cite{Ochs2012} used affinities defined on higher order tuples, which results in a hyper-graph. Using a nonlinear projection this hyper-graph was then converted to an ordinary graph which was segmented using spectral clustering. 

In this analysis we use the approach proposed by Ochs and Brox \cite{Ochs2012} where a motion of an object is modeled using a special similarity transform $\mathcal{T} \in$ SSim(2), with parameters scaling ($s$), rotation ($\alpha$) and translation ($v$). The distance from a trajectory ($c_i(t) \to c_i(t')$) to the model $\mathcal{T}_t$ is calculated using $L_2$-distance $d_{\mathcal{T}_t, i} = \left \| \mathcal{T}_t c_i(t) - c_i(t') \right \|$. A motion hypothesis $\mathcal{T}_t$ at time $t$ can be obtained using two or more point trajectories that exist in the interval $[t,t']$ . In our implementation we used edges of size $h=p+2=4$ to generate hypotheses. It should be noted here that the distance measure is only valid if the trajectories used to generate the hypothesis and the trajectory to which the distance is calculated all coexist in time. Hence a distance of infinity is assigned to all the points that does not exist in the time interval $[t,t']$.  This behavior causes complications in the weight update of the proposed method as now some trajectories can be identified as outliers even though those belong to the same object. To overcome this we uniformly sample small windows (of size 7 frames) and limit the weight updates to that window alone. 

Another important feature of this dataset is that most sequences have a large number of frames and data points (e.g. sequence "tennis" even with 8 times down-scaling \cite{Ochs2012}, includes more than 450 frames and 40,000 data points). Storing a graph of that size is challenging specially on a PC. Hence, in cases where the number of frames is large, we divide the video into few large windows (e.g. 100 frames) and solve the problem in each large window independently. Next we calculated the  mutual distance between each structure in different windows and clustered them using k-means to get the desired number of structures. The number of clusters is a parameter selected such that it would result in reasonable accuracy with least over-segmentation.

Once the clustering was obtained they were evaluated using the method provided along with the dataset (man made masks on specific frames of the videos). We compare our results with \cite{Ochs2012}, \cite{Purkait2014}, which are based on higher order affinities. The results given in \tabref{longtermVid} show that our method has achieved similar accuracies to those with significant improvements in computation time. 
The computation time is related to the number of hyper-edges used and OB used $N^2 \times (30+12)$ hyper edges in their implementation where as HOSC used ${2N}/{5} + N$. In contrast our method uses fewer hyper-edges ($N/10$) selected using the k-th order cost function. The results show that if the edges are selected appropriately similar accuracies can be achieved and lower number of edges means a lower computational time.
We also note here that while the two competing methods \cite{Ochs2012},\cite{Purkait2014} use spacial contiguity in selecting the edges to construct the affinity graph, the proposed method have not used any such additional information.



\begin{table*}
	\caption{Motion segmentation results on Berkeley Motion Segmentation Dataset (BMS-26). }
	\begin{center}
		\footnotesize
		\begin{tabular}{ccccccc}
			\hline
			& Density & Overall error & Average error& Over-segmentation rate & Extracted objects & Total Time(s)\\ \hline 
			OB & 1.03\% & 5.68\%	 & 24.74\% & 1.48 & 30 & 434545\\ 
			HOSC & 1.03\% & 8.05\% & 27.84\% & 2.1  & 22 & 11966\\  
			CBS & 1.03\% & 7.80\% & 22.60\% & 2.08  & 22 & 7875\\  \hline
		\end{tabular}
	\end{center}
	\label{tab:longtermVid}
\end{table*}

\section{Discussion}
\label{sec:Discussion}


The proposed method requires the value of $k$, which defines the minimal acceptable size for a structure in a given application, as an input. Any robust model fitting method needs to establish the minimal acceptable structure size (either explicitly or implicitly), or else it may result in a trivial solution. For example if we are given a set of 2D points and asked to identify lines in data without any additional constraint, there would be no basis to exclude the trivial solution because any two points will result in a perfect line. Hence, in order to find a meaningful solutions there must be some additional constraints such as the minimal acceptable size for a structure. The proposed method estimates the scale of noise from  data and the analysis of \cite{Hoseinnezhad2010} showed that the estimation of the noise scale from data requires at least around 20 data points to limit the effects of finite sample bias. This leads to a lower bound of $k$ around $20$.

Similar to competing clustering based methods (e.g. SCC \cite{Chen2009}, SSC \cite{Elhamifar2013}) the proposed method also requires prior knowledge on the number of clusters. This is one of the limitations of the proposed method. The problem of identifying the number of structures and the scale of noise simultaneously is still a highly researched area. Remaining outliers can always be seen as members of a model with large noise values. Zelnik-Manor and Perona \cite{Zelnik-Manor2004} proposed a method to automatically estimate the number of clusters in a graph using Eigenvector analysis. Since our focus in this paper is on efficiently generating the graph (not in how to cluster it), we have not included this in the evaluations.    
Some model fitting methods that are based on energy minimization \cite{Boykov2001} are devised to estimate the number of structures given the scale of noise. They achieve this by adding a model complexity term to the cost function that penalize additional structures in a given solution. However, these methods require an additional parameter that balances the data fidelity cost with the model complexity (number of structures in \cite{Purkait2014}). Our experiments on \cite{Purkait2014} showed that the output of these methods were heavily dependent on this parameter and required hand tunning on each image (of \tabref{fundamentalRes}) to generate reliable results. 

The proposed method uses a data-sub-sampling strategy based on a set of inclusion weights to bias the algorithm to produce edges from different structures. These inclusion weights iteratively calculated using the inlier/outlier dichotomy for each edge. However in case there are additional information about the problem such as spacial contiguity, one can use those to improve the sub-sampling. For example in two-view motion segmentation, the euclidean distance between points can be used to construct a KDtree, which can then be used to do the sampling directly (i.e. select initial point randomly and include $N_s$ points closest to that point as the data sub-sample). It is important to note that in the performance evaluations of this paper we have not used any such additional information.

\section{Conclusion}
\label{sec:conclusion}

In this paper we proposed an efficient sampling method to obtain a highly accurate approximation of the full graph required to solve the multi-structural model fitting problems in computer vision. The proposed method is based on the observation that the usefulness of a graph for segmentation improves as the distribution of hypotheses (used to build the graph) approaches the actual parameter distribution for the given data. In this paper we approximate this actual parameter distribution using the $k$-th order statistics cost function and the samples are generated using a greedy algorithm coupled with a data sub-sampling strategy. 

The performance of the algorithm in terms of accuracy and computational efficiency was evaluated on several instances of the multi-object motion segmentation problems and was compared with state-of-the-art model fitting techniques. The comparisons show that the proposed method is both highly accurate and computationally efficient.


%

%

\ifCLASSOPTIONcompsoc
  \section*{Acknowledgments}
\else
  \section*{Acknowledgment}
\fi

This research was partly supported under Australian Research Council (ARC) Linkage Projects funding scheme.

\ifCLASSOPTIONcaptionsoff
  \newpage
\fi



\bibliographystyle{IEEEtran}
\end{document}